\documentclass[10pt,journal,compsoc]{IEEEtran}
\ifCLASSOPTIONcompsoc
  \usepackage[nocompress]{cite}
\else
  \usepackage{cite}
\fi
\ifCLASSINFOpdf
\else
\fi

\usepackage{amsmath,amsfonts,bm}

\def\eqref#1{equation~\ref{#1}}

\def\1{\bm{1}}

\DeclareMathAlphabet{\mathsfit}{\encodingdefault}{\sfdefault}{m}{sl}
\SetMathAlphabet{\mathsfit}{bold}{\encodingdefault}{\sfdefault}{bx}{n}

\newcommand{\E}{\mathbb{E}}

\usepackage{booktabs}       
\usepackage{amsfonts}       
\usepackage{nicefrac}       
\usepackage{microtype}      
\usepackage{pgfplots}
\usepackage{algpseudocode}
\usepackage{algorithm}
\usepackage{mathtools}
\usepackage{color}
\usepackage{multirow,graphicx,paralist,bigdelim}
\usepackage{colortbl}
\usepackage{tabularx}
\usepackage{comment}

\usepackage{epstopdf}
\usepackage{epsfig}
\usepackage{graphicx}
\usepackage{amsmath}
\usepackage{amssymb}
\usepackage{pdfrender}
\usepackage{wrapfig}
\usepackage{tikz}
\usepackage{enumitem}

\def\checkmark{\tikz\fill[scale=0.37](0,.35) -- (.25,0) -- (1,.7) -- (.25,.15) -- cycle;}

\usepackage[letterpaper=true,colorlinks=true, linkbordercolor =red, urlcolor=black, bookmarks=false]{hyperref}

\begin{document}
%
\title{Position, Padding and Predictions: \\ A Deeper Look at Position Information in CNNs}
%
%
%
%

\author{Md Amirul Islam,
        Matthew Kowal,
        Sen Jia,
        Konstantinos G. Derpanis,
        and Neil D. B. Bruce
\IEEEcompsocitemizethanks{\IEEEcompsocthanksitem M. A. Islam, M. Kowal, K. G. Derpanis are with Ryerson University, Canada. Email: \{mdamirul.islam, matthew.kowal, kosta\}@ryerson.ca
\IEEEcompsocthanksitem S. Jia is with University of Waterloo, Canada. Email: sen.jia@uwaterloo.ca
\IEEEcompsocthanksitem N. Bruce is with University of Guelph, Canada. Email: brucen@uoguelph.ca
\IEEEcompsocthanksitem M. A. Islam, K. G. Derpanis, and N. Bruce are also with Vector Institute of Artificial Intelligence, Toronto, Canada.
\IEEEcompsocthanksitem K. G. Derpanis is also with Samsung AI Centre Toronto, Canada.
}}
\IEEEtitleabstractindextext{%
\begin{abstract}

In contrast to fully connected networks, Convolutional Neural Networks (CNNs) achieve efficiency by learning weights associated with local filters with a finite spatial extent. An implication of this is that a filter may know what it is looking at, but not where it is positioned in the image. In this paper, we first test this hypothesis and reveal that a surprising degree of absolute position information is encoded in commonly used CNNs. We show that zero padding drives CNNs to encode position information in their internal representations, while a lack of padding precludes position encoding. This gives rise to deeper questions about the role of position information in CNNs: (i) What boundary heuristics enable optimal position encoding for downstream tasks?; (ii) Does position encoding affect the learning of semantic representations?; (iii) Does position encoding always improve performance? To provide answers, we perform the largest case study to date on the role that padding and border heuristics play in CNNs. We design novel tasks which allow us to quantify boundary effects as a function of the distance to the border. Numerous semantic objectives reveal the effect of the border on semantic representations. Finally, we demonstrate the implications of these findings on multiple real-world tasks to show that position information can both help or hurt performance.

\end{abstract}

}

\maketitle

\IEEEdisplaynontitleabstractindextext

%
\IEEEpeerreviewmaketitle

\section{Introduction}
\begin{figure}
	\begin{center}
	\setlength\tabcolsep{2.7pt}
	\def\arraystretch{0.5}
	\resizebox{0.45\textwidth}{!}{
		\begin{tabular}{*{2}{c }}
			\includegraphics[width=0.25\textwidth,  height=2cm]{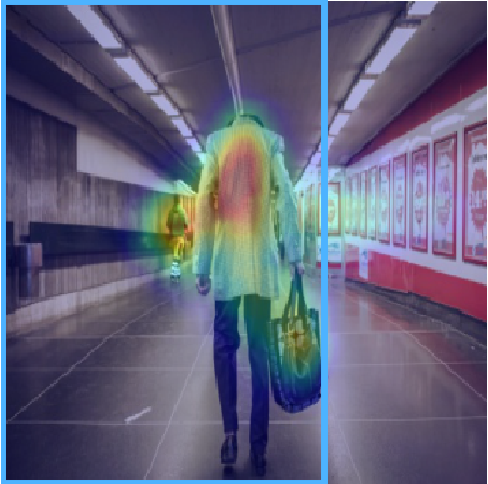}&
			\includegraphics[width=0.17\textwidth,  height=2cm]{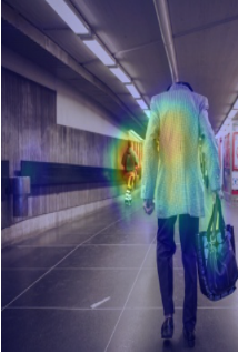} \\
			\includegraphics[width=0.25\textwidth,  height=2cm]{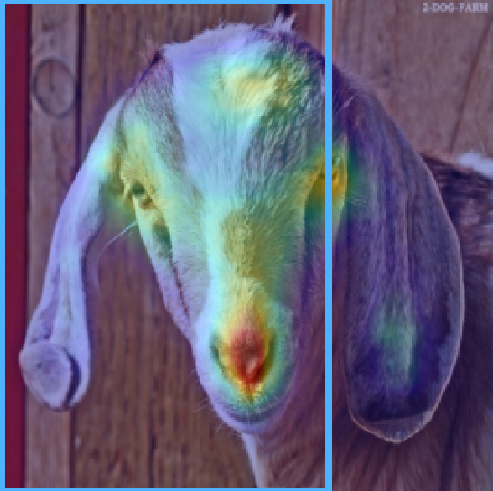}&
			\includegraphics[width=0.17\textwidth,  height=2cm]{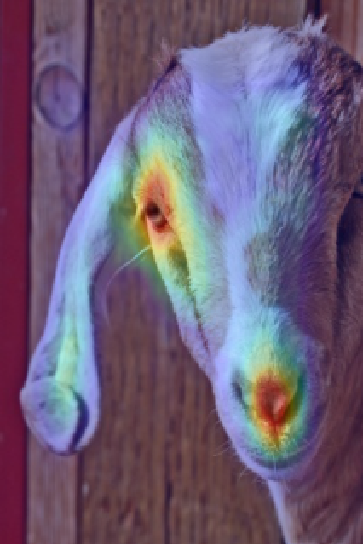} \\
			
			\includegraphics[width=0.25\textwidth,  height=2cm]{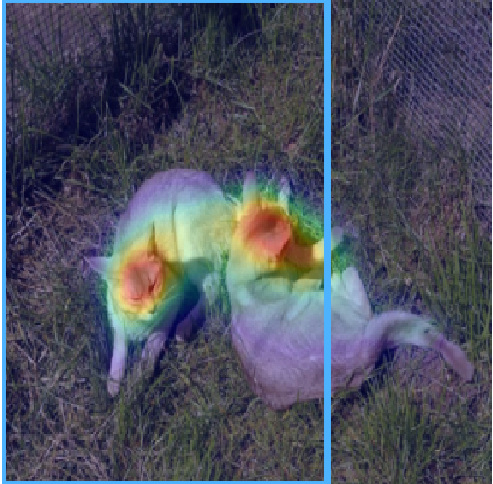}&
			\includegraphics[width=0.17\textwidth,  height=2cm]{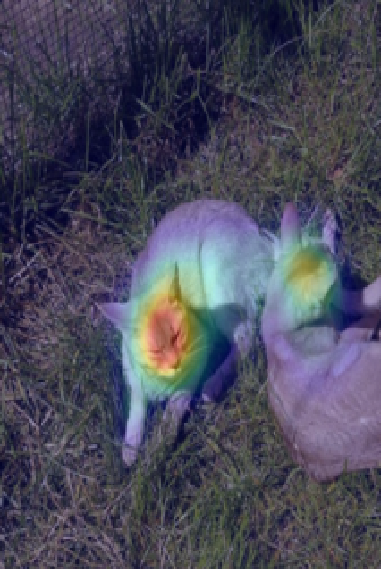}\\
			Image & Cropped Image
		\end{tabular}}
		\vspace{-0.1cm}
		\caption{Sample predictions for salient regions for input images (left), and a slightly cropped version (right). Cropping results in a shift in position rightward of features relative to the centre. It is notable that this has a significant impact on output and decision of regions deemed salient despite no explicit position encoding and a modest change to position in the input.}
		\label{fig:intro_1}
		\end{center}
		\vspace{-0.5cm}
\end{figure}

One of the main intuitions behind the success of CNNs for visual tasks such as image classification~\cite{krizhevsky2012imagenet,Simonyan14,szegedy2015going,huang2017densely}, video classification~\cite{karpathy2014large,yue2015beyond,carreira2017quo}, object detection~\cite{Ren15,Redmon15,he2017mask}, generative image models~\cite{brock2018large}, semantic segmentation~\cite{long15_cvpr,noh15_iccv,islam2017label,islam2017gated,chen2017rethinking,islam2018gated,karim2019recurrent,chen2018deeplab,karim2020distributed, islam2020sbinding}, and saliency detection~\cite{Liu_2016_CVPR,islam2017salient,cvpr18_rank,PICA,islam2018semantics,kalash2019relative}, is that convolutions are translation equivariant. This adds a visual inductive bias to the neural network which assumes that objects can appear \textit{anywhere in the image}. Thus, CNNs are considered to be spatially agnostic. 
However, until recently, it was unclear if CNNs encode any absolute spatial information, which may be important for tasks that are dependant on the position of objects in the image (e.g., semantic segmentation and salient object detection). For example, while detecting saliency on a cropped version of the images, the most salient region shifts even though the visual features have not been changed. As shown in Fig.~\ref{fig:intro_1}, the regions determined to be most salient~\cite{Jia18} tend to be near the \textit{center} of an image. This is somewhat surprising, given the limited spatial extent of CNN filters through which the image is interpreted. 
In this paper, we first examine the role of absolute position information by performing a series of \textit{randomization tests} with the hypothesis that CNNs might indeed learn to encode position information as a cue for decision making. Our experiments reveal that position information is implicitly learned from the commonly used padding operation (\textit{zero padding}). Padding is commonly used to accommodate the finite domain of images and to allow the convolutional kernel's support to extend beyond the border of an image and reduce the impact of the \textit{boundary effects}~\cite{wohlberg2017convolutional,tang2018high,liu2018partial,innamorati2019learning,liu2018intriguing}. In particular, zero padding is widely used for keeping the same dimensionality when applying convolution. However, its hidden effect in representation learning has long been ignored. 

\begin{figure}
  \begin{center}
      \includegraphics[width=0.48\textwidth]{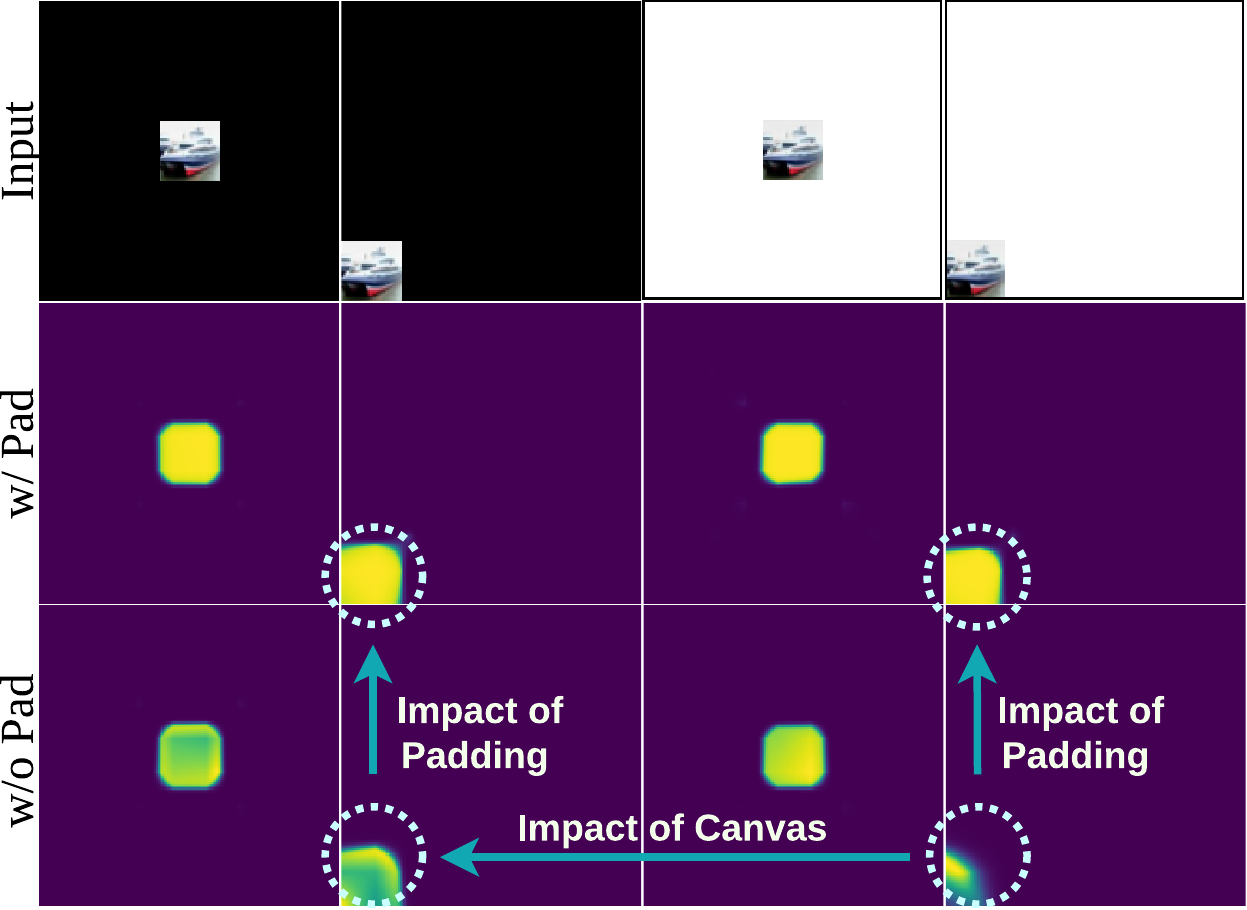} 
  \end{center}
  \vspace{-0.25cm}
\caption{An illustration of how border color and padding changes the boundary effects. We place CIFAR-10 images in random locations on a canvas of $0$'s (black) or $1$'s (white). We evaluate if a ResNet-18, trained w/ or w/o padding for semantic segmentation, can segment the image region. Surprisingly, performance is improved when either zero padding or a black canvas is used, implying position information can be exploited from border heuristics to reduce the boundary effect. Colormap is `viridis'; yellow is high confidence.}
  \label{fig:intro}
  \vspace{-0.45cm}
\end{figure}

Recent studies~\cite{perez2019turing,kayhan2020translation} also have shown that zero padding allows CNNs to encode absolute position information despite the presence of pooling layers in their architecture (e.g., global average pooling). In our work, we argue that the relationship between \textit{boundary effects} and absolute position information extends beyond zero padding and has major implications in a CNN's ability to encode confident and accurate semantic representations (see Fig.~\ref{fig:intro}). Our work helps to better understand the nature of the learned features in CNNs, with respect to the interaction between padding usage and positional encoding, and highlights important observations and fruitful directions for future investigation.

Another unexplored area related to boundary effects is the use of \textit{canvases} (i.e., backgrounds) with image patches (see Fig.~\ref{fig:intro}, top row). When using image patches in a deep learning pipeline involving CNNs, the user is required to paste the patch onto a canvas due to the constraint that the image must be rectangular. Canvases have been used in a wide variety of domains, such as image generation~\cite{gregor2015draw, huang2019learning}, data augmentation~\cite{devries2017improved}, image inpainting~\cite{demir2018patch,yu2018generative}, and interpretable AI~\cite{geirhos2018imagenet,esser2020disentangling}. To the best of our knowledge, we first analyze the relationship between canvas value selection and absolute position information.
In other works, the canvas value is simply chosen in an adhoc manner, without consideration to the possible downstream implications. 

Given the pervasiveness of CNNs in a multitude of applications, it is of paramount importance to fully understand what the internal representations are encoding in these networks, as well as isolating the precise reasons that these representations are learned. This comprehension can also allow for the effective design of architectures that overcome recognized shortcomings (e.g., residual connections~\cite{he2016deep} for the vanishing gradient problem). As boundary effects and position information in CNNs are still largely not fully understood, we aim to provide answers to the following hypotheses which reveal fundamental properties of these phenomena: 

\noindent \textbf{Hypothesis I: Zero Padding Encodes Maximal Absolute Position Information:} Does zero padding encode maximal position information compared to other padding types? We evaluate the amount of position information in networks trained with different padding types and show zero padding injects more position information than common padding types such as reflection, replicate, and circular. 

\noindent \textbf{Hypothesis II: Different Canvas Colors Affect Performance:} Do different background values have an effect on performance? If the padding value at the boundary has a substantial effect on a CNNs performance and position information contained in the network, one should expect that canvas values may also have a similar effect. 

\noindent \textbf{Hypothesis III: Position information is Correlated with Semantic Information:} Does a network's ability to encode absolute position information affect its ability to encode semantic information? If zero padding and certain canvas colors can affect performance on classification tasks due to increased position information, we expect that the position information is correlated with a network's ability to encode semantic information. We demonstrate that encoding position information improves the robustness and separability of semantic features.

\noindent \textbf{Hypothesis IV: Boundary Effects Occur at All Image Locations:} Does a CNN trained without padding suffer in performance solely at the border, or at all image regions? How does the performance change across image locations? Our analysis reveals strong evidence that the border effect impacts a CNN's performance at \textit{all regions} in the input, contrasting previous assumptions~\cite{tsotsos1995modeling,innamorati2019learning} that border effects exist solely at the image border.

\noindent \textbf{Hypothesis V: Position Encoding Can Act as a Feature or a Bug:} Does absolute position information always correlate with improved performance? A CNN’s ability to leverage position information from boundary information could hurt performance when a task requires translation-invariance, e.g., texture recognition; however, it can also be useful if the task relies on position information, e.g., semantic segmentation.

To give answers to these hypotheses (hereon referred to as \textbf{H-X}), we design a series of novel tasks as well as use existing techniques to quantify the absolute location information contained in different CNNs with various settings. The contribution of this paper extends from the analysis presented in our prior work~\cite{islam2020much} which demonstrates that (i) CNNs encode absolute position information and (ii) zero padding is a main source of this positional information in CNNs. We extend our prior work in the following respects:

\begin{itemize}
    \setlength{\parskip}{0pt}
    \setlength{\parsep}{0pt}
    \item We introduce location dependant experiments (see Fig.~\ref{fig:spectrum}) which use a grid-based strategy to allow for a  per-location analysis of border effects in relation to absolute position information. We demonstrate that the per-location analysis plays a crucial role in determining the isolated impact between boundary effects and absolute position information as a function of the distance to the image border.
    \item We show zero padding implicitly injects more position information than common padding types (e.g., reflection, replicate, and circular).
    
    \item  We estimate the number of dimensions which encode position information in the latent representations of CNNs. 
    
    \item Through these experiments we show both quantitative and qualitative evidence that boundary effects have a substantial effect on CNNs in surprising ways and then demonstrate the practical implications of these findings on multiple real-world applications. Code will be made available for all experiments.
\end{itemize}

\section{Related Work}
\noindent \textbf{Absolute Position Information in CNNs.} Many studies have explored various mechanisms which allow for humans to understand the learning process of CNNs, e.g., visualization of features~\cite{Zeiler11, ZeilerF13}, understanding generalization~\cite{Zhang16}, Class Activation Maps (CAMs)~\cite{zhou2016learning,Selvaraju16}, and disentangling representations~\cite{denton2017unsupervised, lorenz2019unsupervised, esser2020disentangling}. Recent works have explored this area in relation to a CNN's ability to encode absolute position information. In particular,~\cite{kayhan2020translation,alsallakh2020mind} have shown that CNNs are able to exploit absolute position information despite the pooling operation. This is consistent with the findings of our prior work~\cite{islam2020much} where we showed that a decoder module can extract pixel-wise location information from the encodings of a CNN. We further suggested that zero padding is a key source of the encoded position information and revealed that a padding of size two enables CNNs to encode more position information.~\cite{kayhan2020translation} also pointed out that a padding size of two enables all pixels in the input to have an equal number of convolution operations performed on it and showed further beneficial properties of this padding type, such as data efficiency. 
~\cite{alsallakh2020mind} observe the similar phenomenon and find that such spatial bias cause blind spots for small object detection.
~\cite{xu2020positional} investigated different positional encodings and analyze their effects in generating images. In contrast, we design novel experiments which allow us to conduct a \textit{distance-to-border} analysis to reveal characteristics of the \textit{relationship} between the boundary effect and a CNN's ability to exploit absolute position information.\\

\noindent \textbf{Explicit Positional Encoding.} Another line of research~\cite{wang2018location,liu2018intriguing,murase2020can} explicitly injects absolute location information with the intuition of exploiting location bias in the network to improve the performance on several tasks. In~\cite{wang2018location}, the input image is augmented with additional location information which improves the performance of the CNN on salient object segmentation and semantic segmentation. Another simple approach to inject location information is introduced in~\cite{liu2018intriguing}, where an additional channel is appended to convolutional layers containing the spatial location of the convolutional filter. Improvements with this layer augmentation are shown on a variety of tasks, including image classification, generative modelling, and object detection. Additionally, various forms of position information have been injected in neural networks through the use of capsule \cite{Sabour17} and recurrent networks~\cite{Visin15}, which encode relative spatial relationships within learned feature layers.\\

\noindent \textbf{Boundary Effects in CNNs.} The boundary effect is a well studied phenomenon in biological neural networks~\cite{sirovich1979effect,tsotsos1995modeling}. Previous works that have considered the boundary effect for artificial CNNs, have done so by means of using specialized convolutional filters for the border regions~\cite{innamorati2019learning}, or re-weighting convolution activations near the image borders by the ratio between the padded area and the convolution window area~\cite{liu2018partial}. 

The groundwork for some of what is presented in this paper  appeared  previously ~\cite{islam2020much}, in which we have shown that CNNs encode absolute position information and zero padding delivers the position information. This give rise to deeper questions about the role of absolute position information to address boundary effects in CNNs. In this work, we specifically focus on the relationship between boundary effects and absolute position information with respect to padding. This is accompanied by an in depth analysis of introduced location dependent tasks with a per-location analysis of border effects.  

\section{Absolute Position Information in CNNs}\label{sec:pad}
In this section, we revisit the hypothesis presented in our prior work~\cite{islam2020much} that position information is implicitly encoded within the extracted feature maps from a pretrained CNNs. We validate this hypothesis empirically by predicting position information from different CNN archetypes in an end-to-end manner. In the following subsections, we first summarize the problem definition, position encoding network, and synthetic data generation. Then we discuss the existence (Sec.~\ref{sec:position}) and source (Sec.~\ref{sec:pos_source}) of position information followed by the comparison of different padding types in terms of encoding position information (Sec.~\ref{sec:pad_com}). \\

\begin{figure}[t]
		\resizebox{0.49\textwidth}{!}{

			\setlength\tabcolsep{1.2pt}
			\def\arraystretch{0.6}
			\begin{tabular}{*{5}{c }}
						
						\includegraphics[width=0.15\textwidth]{imgs/sample/gt_hor}&
						\includegraphics[width=0.15\textwidth]{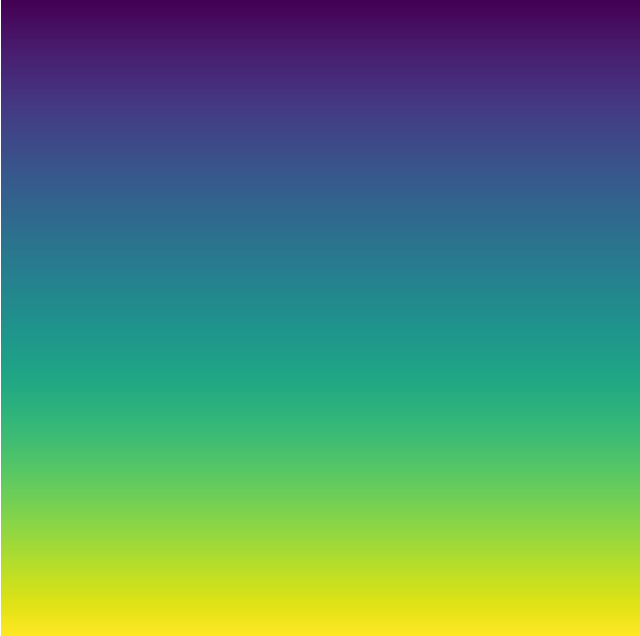}&
						\includegraphics[width=0.15\textwidth]{imgs/sample/gt_gau}&
						\includegraphics[width=0.15\textwidth]{imgs/sample/gt_horstp}&
						\includegraphics[width=0.15\textwidth]{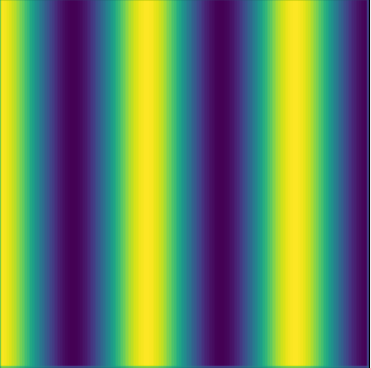}\\
						
						 H &  V &  G & HS & VS
					\end{tabular}
					
				}
		\vspace{-0.1cm}
		\caption{Generated gradient-like ground-truth position maps. \textbf{H}: Horizontal, \textbf{V}: Vertical, \textbf{G}: Gaussian, \textbf{HS}: Horizontal Stripe, \textbf{VS}: Vertical Stripe.}
		\label{fig:gtmaps}
		\vspace{-0.1cm}
\end{figure}
\noindent \textbf{Problem Formulation.}
Given an input image $\mathcal{I}_m{\in\mathbb{R}^{h\times w \times 3}}$, our goal is to predict a gradient-like position information mask, $f_p{\in\mathbb{R}^{h\times w}}$, where each pixel value defines the absolute coordinates of a pixel from left$\rightarrow$right or top$\rightarrow$bottom. We generate gradient-like masks, $\mathcal{G}_{p}{\in\mathbb{R}^{h\times w}}$, for supervision in our experiments, with weights of the base CNN archetypes being fixed.\\

\noindent \textbf{Position Encoding Network.}
Our \textit{Position Encoding Network} (PosENet) consists of two key components: a feed-forward convolutional encoder network and a simple position encoding module (PosEnc). The encoder network extracts features at different levels of abstraction, from shallower to deeper layers. The position encoding module takes multi-scale features from the encoder network as input and predicts the absolute position information.\\

\noindent \textbf{Synthetic Data and Ground-truth Generation.}\label{sec:synthetic}
To validate the existence of position information in a network, we implement a \textit{randomization test} by assigning a normalized gradient-like 
\footnote{We use the term \textit{gradient} to denote pixel intensities instead of the gradient in back propagation.} 
position map as ground-truth shown in Fig.~\ref{fig:gtmaps}. We first generate gradient-like masks in Horizontal (\textbf{H}) and Vertical (\textbf{V}) directions. Similarly, we apply a Gaussian filter to design another type of ground-truth map, Gaussian distribution (\textbf{G}). The key motivation of generating these three patterns is to validate if the model can learn absolute position on one or two axes. Additionally, we also create two types of repeated patterns, horizontal and vertical stripes, (\textbf{HS}, \textbf{VS}). Regardless of the direction, the position information in the multi-level features is likely to be modelled through a transformation by the encoding module. Our design of gradient ground-truth can be considered as a type of random label because there is no correlation between the input image and the ground-truth with respect to position. Since the extraction of position information is independent of the content of images, we can choose any image datasets. Meanwhile, we also build synthetic images (\textit{Black} and \textit{White}) to validate our hypothesis.

\begin{table}
\caption{Quantitative comparison of different PosENets in terms of SPC and MAE across various image types. VGG and ResNet based PosENet can decode absolute position information more easily compared to the PosENet without any backbone network.}
	    \centering
			\setlength\tabcolsep{3.7pt}
		    \def\arraystretch{1.15}
			\resizebox{0.49\textwidth}{!}{
				\begin{tabular}{c|l|cc|cc|cc}
					\specialrule{1.2pt}{1pt}{1pt}\
					&\multirow{2}{*}{Model}&\multicolumn{2}{c|}{PASCAL-S}& \multicolumn{2}{c|}{Black}& \multicolumn{2}{c}{White}\\
					\cline{3-8}
					&& SPC$\uparrow$ & MAE$\downarrow$ & SPC$\uparrow$ & MAE$\downarrow$ & SPC$\uparrow$ & MAE$\downarrow$\\
					\specialrule{1.2pt}{1pt}{1pt}\
					
					\multirow{3}{*}{{\textbf{H}}}&
					PosENet & .01 & .25 & .0&.25&.0&.25\\
					&VGG & .74&.15 & .75&.16&.87&.16\\
					&ResNet  & .93&.08& .99&.08&.99&.08\\
					
					\hline
					\multirow{3}{*}{{\textbf{V}}}&
					PosENet & .13 & .25& .0&.25&.0&.25\\
					&VGG  & .82 & .13 & .85 & .15 & .93 & .14 \\
					&ResNet  & .95&.08& .98&.07&.98&.07\\
					
					\hline
					\multirow{3}{*}{{\textbf{G}}}&
					PosENet & -.01 & .23& .0&.19&.0&.19\\
					&VGG & .81 & .11 & .84 & .12 & .90 & .12 \\
					&ResNet  & .94& .07 & .95 & .07 & .96 &.06\\
					
					\hline
					\multirow{3}{*}{{\textbf{HS}}}&
					PosENet & -.01 & .71& -.06&.70&.0&.70\\
					&VGG & .41 & .56 & .53 & .58 & .58 & .57 \\
					&ResNet  & .53 & .53 & .57 & .52 & .56 & .52\\
					
					\hline
					\multirow{3}{*}{{\textbf{VS}}}&
					PosENet & .01 & .72& .08 & .71&.08&.71\\
					&VGG & .37 & .57 & .54 & .58 & .44 & .58 \\
					&ResNet  & .52 & .54 & .57 & .52 & .59 & .51 \\
					\specialrule{1.2pt}{1pt}{1pt}
				\end{tabular}
			}
			
			\label{tab:overall_result}
\end{table}	
\begin{figure}
	    \centering
			
			\resizebox{0.49\textwidth}{!}{
				\setlength\tabcolsep{1.2pt}
				\def\arraystretch{0.7}
				\begin{tabular}{*{5}{c }}
					\includegraphics[width=0.11\textwidth,height=1.8cm]{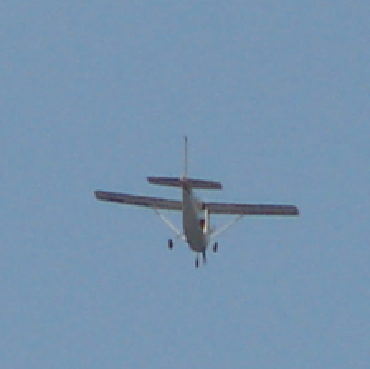}&
					\includegraphics[width=0.11\textwidth,height=1.8cm]{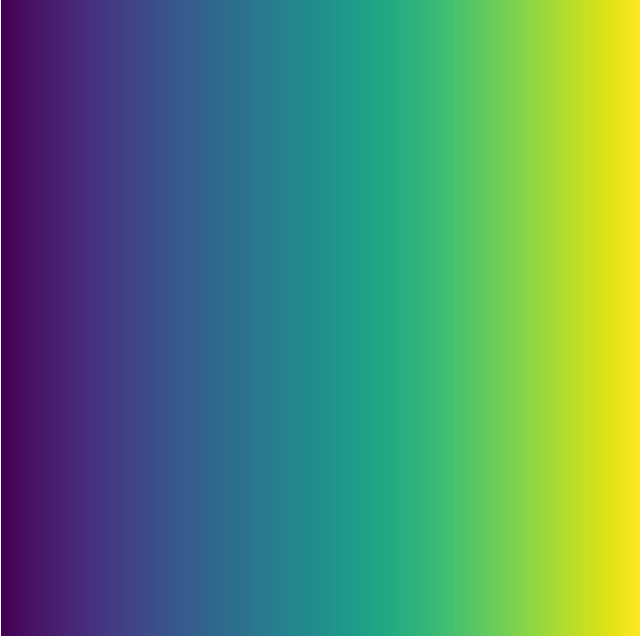}&
					\includegraphics[width=0.11\textwidth,height=1.8cm]{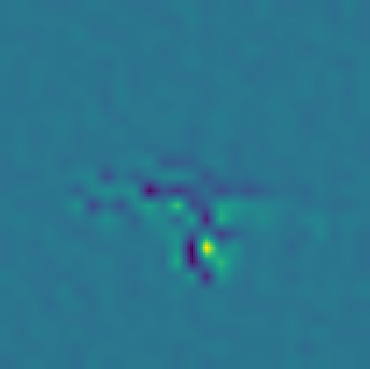}&
					\includegraphics[width=0.11\textwidth,height=1.8cm]{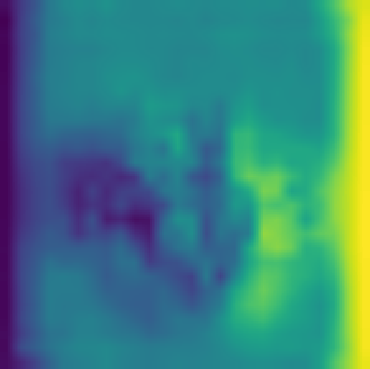}&
					\includegraphics[width=0.11\textwidth,height=1.8cm]{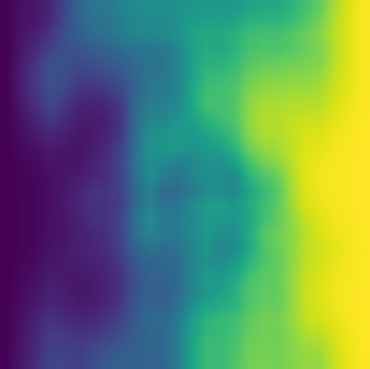} \\
					
					\includegraphics[width=0.11\textwidth,height=1.8cm]{imgs/result/img.png}&
					\includegraphics[width=0.11\textwidth,height=1.8cm]{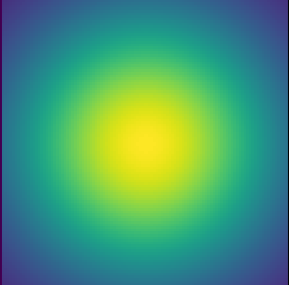}&
					\includegraphics[width=0.11\textwidth,height=1.8cm]{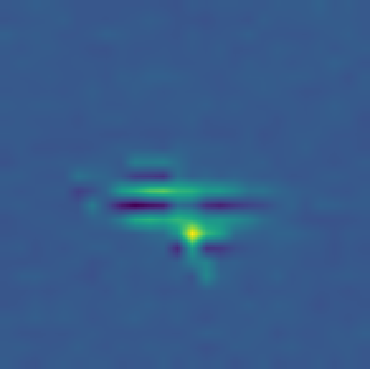}&
					\includegraphics[width=0.11\textwidth,height=1.8cm]{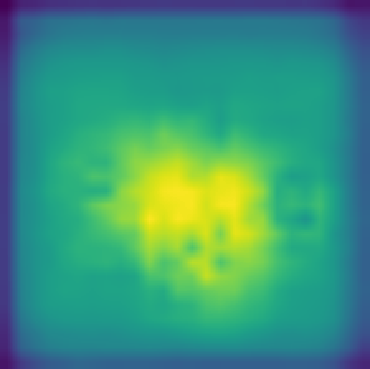}&
					\includegraphics[width=0.11\textwidth,height=1.8cm]{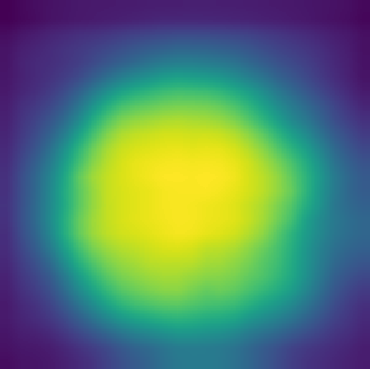} \\
					
					\includegraphics[width=0.11\textwidth,height=1.8cm]{imgs/result/img.png}&
					\includegraphics[width=0.11\textwidth,height=1.8cm]{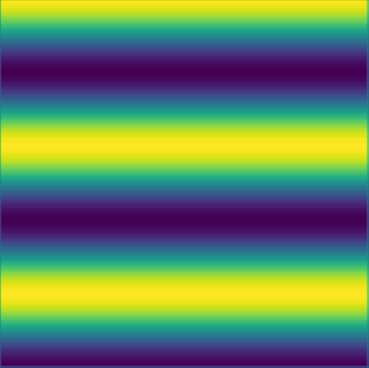}&
					\includegraphics[width=0.11\textwidth,height=1.8cm]{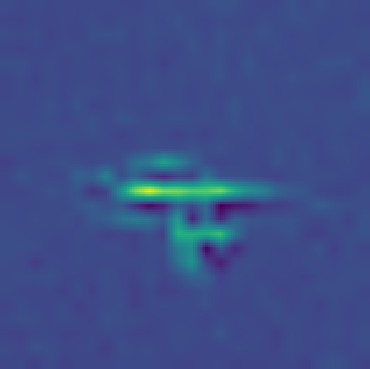}&
					\includegraphics[width=0.11\textwidth,height=1.8cm]{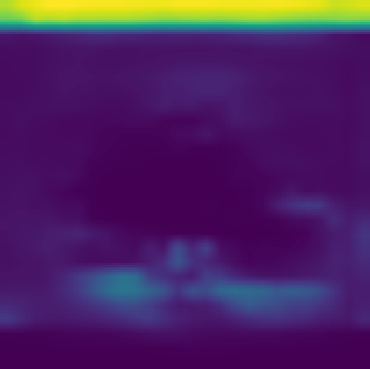}&
					\includegraphics[width=0.11\textwidth,height=1.8cm]{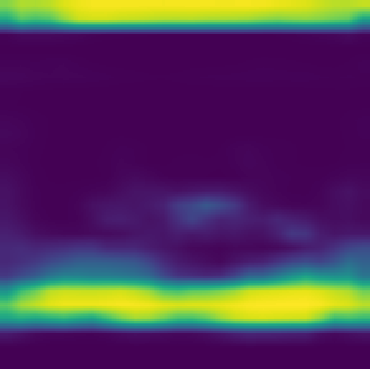} \\
					
				 Input &  GT &  PosENet & VGG & ResNet 
						
				\end{tabular}
			}
			\vspace{-0.1cm}
			\caption{Qualitative results of PosENet based networks corresponding to different ground-truth patterns.}
			\label{fig:val_pascal}
\end{figure}
\subsection{Existence of Position Information}\label{sec:position}
We first conduct experiments to validate the existence of position information encoded in a pretrained CNN model. We report experimental results for the following baselines that are described as follows: \textbf{VGG} indicates PosENet is based on the features extracted from the VGG16 model. Similarly, \textbf{ResNet} represents the combination of ResNet-152 and PosENet. \textbf{PosENet} alone denotes only the PosENet model is applied to learn position information directly from the input image. Following the experimental details provided in Appendix A.1, we train the VGG16~\cite{Simonyan14} and ResNet152~\cite{he2016deep} based PosENet on each type of the ground-truth and report the experimental results in Table~\ref{tab:overall_result}. We also report results when we only train PosENet without using any pretrained model to justify that the position information is not driven from prior knowledge of objects. Our experiments do not focus on achieving higher performance on the metrics but instead validate how much position information a CNN model encodes or how easily PosENet can extract this information. Note that, we only use one convolutional layer with a kernel size of $3 \times 3$ without any padding in the position encoding module for this experiment.

As shown in Table~\ref{tab:overall_result}, PosENet (VGG16 and ResNet152) can easily extract absolute position information from the pretrained CNN models, especially the ResNet152 based PosENet model. However, training the position encoding module (PosENet in Table~\ref{tab:overall_result}) without any pretrained encoder achieves much lower scores across different patterns and source images. This result implies that it is very difficult to extract position information from the input image alone. PosENet can extract position information consistent with the ground-truth position map only when coupled with a deep encoder network. As mentioned prior, the generated ground-truth map can be considered as a type of \textit{randomization test} given that the correlation with input has been ignored \cite{Zhang16}. Nevertheless, the high performance on the test sets across different ground-truth patterns reveals that the model is not blindly overfitting to the noise and instead is extracting true position information. However, we observe low performance on the repeated patterns (\textbf{HS} and \textbf{VS}) compared to other patterns due to the model complexity and specifically the lack of correlation between ground-truth and absolute position (last two set rows of Table~\ref{tab:overall_result}). The \textit{H} pattern can be seen as one quarter of a sine wave whereas the striped patterns (\textit{HS} and \textit{VS}) can be considered as repeated periods of a sine wave which requires a deeper comprehension.

The qualitative results for several architectures across different patterns are shown in Fig.~\ref{fig:val_pascal}. We can see the correlation between the predicted and the ground-truth position maps corresponding to \textbf{H}, \textbf{G}, and \textbf{HS} patterns, which further reveals the existence of absolute position information in these networks. The quantitative and qualitative results strongly validate our hypothesis that position information is implicitly encoded in every architecture without any explicit supervision towards this objective.

Moreover, PosENet without any backbone encoder shows no capacity to output a gradient map based on the synthetic data. We explored the effect of image semantics in our prior work~\cite{islam2020much}. It is interesting to note the performance gap among different architectures specifically the ResNet based models achieve higher performance than the VGG16 based models. The reason behind this could be the use of different convolutional kernels in the architecture or the degree of prior knowledge of the semantic content. 
\subsection{Where is the Position Information Stored?}
Our previous experiment reveal that the position information is encoded in a pretrained CNN model. It is also interesting to see whether position information is equally distributed across the stages of the pretrained CNN model. In this experiment, we train VGG16 based PosENet on the extracted features of all the stages, $f_1$, $f_2$, $f_3$, $f_4$, $f_5$ separately using VGG16 to examine which layer encodes more position information. Similar to Sec.~\ref{sec:position}, we only apply one $3 \times 3$ kernel in the position encoding module to obtain the position map.

\begin{table}
\caption{Performance of VGG-16~\cite{Simonyan14} on PASCAL-S images with a varying extent of the reach of different feed-forward blocks. Deeper layers in CNNs contain more absolute position information than earlier layers.}
\vspace{-0.1cm}
\def\arraystretch{1.1}
\centering
\resizebox{0.48\textwidth}{!}{
	\begin{tabular}{c|ccccc|c|c}
	\specialrule{1.2pt}{1pt}{1pt}
	&  $f_{1}$ & $f_2$ & $f_3$ & $f_4$  &$f_5$& SPC$\uparrow$ & MAE$\downarrow$    \\
	\specialrule{1.2pt}{1pt}{1pt}
	
	\multirow{5}{*}{{\textbf{H}}}
	&$\checkmark$ &&&&& .101& .249\\
	&&$\checkmark$&&&& .344& .225\\
	& &&$\checkmark$&&& .472& .203\\
	& &&&$\checkmark$&& .610& .181\\
	& &&&&$\checkmark$& .657& .177\\
	& $\checkmark$&$\checkmark$&$\checkmark$&$\checkmark$&$\checkmark$ & .742 & .149\\
	
	\hline
	\multirow{5}{*}{{\textbf{G}}}
	&$\checkmark$ &&&&& .241 & .182\\
	&&$\checkmark$ &&&& .404& .168\\
	& &&$\checkmark$&&& .588 & .146\\
	& &&&$\checkmark$&& .653& .138\\
	& &&&& $\checkmark$& .693& .135\\
	& $\checkmark$&$\checkmark$&$\checkmark$&$\checkmark$&$\checkmark$& .814 & .109\\
	
	\specialrule{1.2pt}{1pt}{1pt}
	\end{tabular}}
    
    \label{tab:feature_vgg16}
\end{table}
As shown in Table~\ref{tab:feature_vgg16}, the VGG based PosENet with $f_5$ features achieves higher performance compared to the $f_1$ features. This may partially a result of more feature maps being extracted from deeper as opposed to shallower layers, $512$ vs $64$ respectively. However, it is likely indicative of stronger encoding of the positional information in the deepest layers of the network where this information is shared by high-level semantics. We further investigate this effect for VGG16 where the top two layers ($f_4$ and $f_5$) have the same number of features. More interestingly, $f_5$ achieves better results than $f_4$. This comparison suggests that the deeper feature contains more position information, which validates the common belief that top level visual features are associated with global features.
\subsection{Where does Position Information Come From?} \label{sec:pos_source}
We hypothesize that the padding near the border delivers a signal which contains positional information. Zero padding is widely used in convolutional layers to maintain the same spatial dimensions for the input and output, with a number of zeros added at the beginning and at the end of both axes, horizontal and vertical. To validate this, we remove all the padding mechanisms implemented within VGG16 but still initialize the model with the ImageNet pretrained weights. Note that we perform this experiment only using VGG16 based PosENet. 
We first test the effect of zero padding used in VGG16, no padding used in the position encoding module. As we can see from Table~\ref{tab:padding_result}, the VGG16 model without zero padding achieves much lower performance than the default setting (padding=1) on the natural images. Similarly, we introduce position information to the PosENet by applying zero padding. PosENet with \textit{padding}=1 (standard zero padding) achieves higher performance than the original (\textit{padding}=0). When we set \textit{padding}=2 (referred as Full-Conv in recent works~\cite{kayhan2020translation,alsallakh2020mind}), the role of position information is more obvious. This also validates our experiment in Section~\ref{sec:position}, that shows PosENet is unable to extract noticeable position information because no padding was applied, and the information is encoded from a pretrained CNN model. This is why we did not apply zero-padding in PosENet in our previous experiments. Moreover, we aim to explore how much position information is encoded in the pretrained model instead of directly combining with the PosENet. 
\begin{table}
\caption{Quantitative comparison subject to padding in the convolution layers used in PosENet and VGG-16~\cite{Simonyan14} (w/o and with zero padding) on PASCAL-S images. The role of position information is more obvious with the increase of padding.}
	\centering
    \resizebox{0.49\textwidth}{!}{
	\def\arraystretch{1.1}
	\centering
	\begin{tabular}{l|cc|cc}
	\specialrule{1.2pt}{1pt}{1pt}\
    \multirow{2}{*}{Model}& \multicolumn{2}{c|}{\textbf{H}} &\multicolumn{2}{c}{\textbf{G}} \\
    \cline{2-5}
    & SPC$\uparrow$  & MAE$\downarrow$  & SPC$\uparrow$  & MAE$\downarrow$ \\
	\specialrule{1.2pt}{1pt}{1pt}
	
	PosENet & .012 &.251 & -.001 & .233 \\
	PosENet w/ \textit{padding}=1 & .274 & .239 & .205 & .184  \\
	PosENet w/ \textit{padding}=2 & .397 &.223& .380 & .177 \\
	VGG16~\cite{Simonyan14} & .742& .149& .814 & .109\\
    VGG16 w/o \textit{padding} & .381 & .223& .359 & .174 \\
	\specialrule{1.2pt}{1pt}{1pt}
	\end{tabular}
    }
    
    \label{tab:padding_result}
\end{table}
\begin{table}[t]
	\caption{Position encoding results with metrics SPC$\uparrow$: high is better and MAE$\downarrow$: low is better, with different padding types. $\dagger$ denotes zero-padding based methods. Zero padding encodes maximal absolute position information compared to other common adding types.} 
	\centering
	\def\arraystretch{1.1}
	\setlength\tabcolsep{3.2pt}
		\resizebox{0.48\textwidth}{!}{
			\begin{tabular}{c|l|cc|cc}
				\specialrule{1.2pt}{1pt}{1pt}
				\multirow{2}{*}{\hspace{0.01cm} $\ast$} &\multirow{2}{*}{\hspace{0.01cm} Padding}& \multicolumn{2}{c|}{\textbf{Horizontal} \hspace{0.1cm}\includegraphics[width=2.9mm, height=2.9mm]{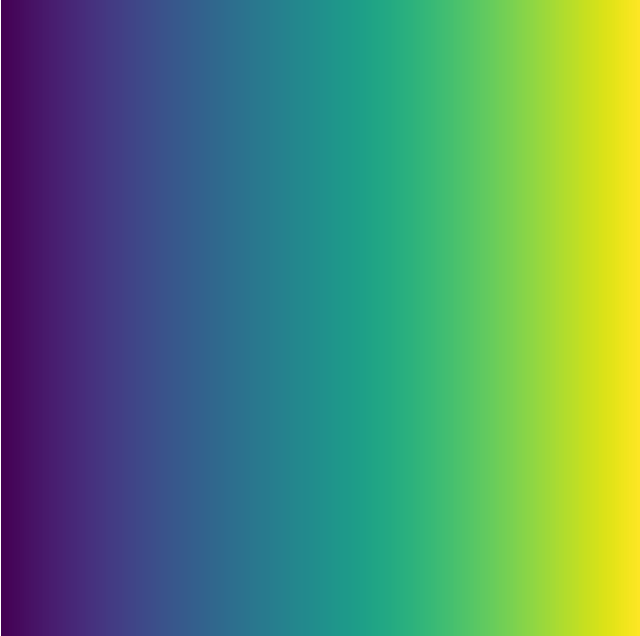}} & 
				\multicolumn{2}{c}{\textbf{Gaussian} \hspace{0.1cm}\includegraphics[width=2.9mm, height=2.9mm]{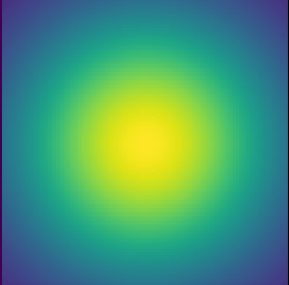}}  \\
				\cline{3-6}
				 &&SPC$\uparrow$ & MAE$\downarrow$ & SPC$\uparrow$ & MAE$\downarrow$ \\
				
				\specialrule{1.2pt}{1pt}{1pt}	
		       
				
				\multirow{6}{*}{\hspace{0.01cm} VGG-5}&Zero Pad$^\dagger$ & .406 & .216 & .591 & .146
				\\
				
				&Partial$^\dagger$ \cite{liu2018partial}   & .424 & .213 & .604 &.144  \\
				
			   
			   &Circular   & .296 & .236 & .455 & .165
			   \\
			   
		       
		       &Replicate   & .218 & .241 & .396 & .173
		       \\
			    
			    
			    &Reflect   & .212 & .242 &.409 & .172
			    \\
			    
			    &w/o Pad & .204 & .243 & .429 & .168
			    \\
			    
				\specialrule{1.2pt}{1pt}{1pt}
			\end{tabular}\label{tab:pad_alg}
	}

    \label{tab:pad_alg_new}
	\vspace{-0.3cm}
\end{table}

\subsection{What Type of Padding Injects Optimal Location Information?} \label{sec:pad_com}
With the ultimate goal of revealing characteristics that determine the impact that \textit{boundary effects} plays in CNNs with respect to absolute position information, we first determine which commonly used padding type encodes the maximum amount of absolute position information. We evaluate the ability of different padding types (i.e., zero, circular, reflection, and replicate) to encode absolute position information by extending the experiments from Sec.~\ref{sec:position}, which only considered zero padding. We first train a simplified VGG classification network~\cite{Simonyan14} with five layers (VGG-5, see Appendix A.2 for implementation details) on Tiny ImageNet~\cite{le2015tiny} for each padding type. We follow the settings as in Sec.~\ref{sec:position}: a position encoding read-out module, trained using DUT-S~\cite{DUTS} images, takes the features from a frozen VGG-5 model's last layer, pre-trained on Tiny ImageNet, and predicts a gradient-like position map (see top row in Table.~\ref{tab:pad_alg_new}). We experiment with two GT position maps, which are the same for every image: (i) `horizontal' and (ii) `Gaussian'. We report results using Spearman Correlation (SPC) and Mean Absolute Error (MAE) with input images from PASCAL-S~\cite{PASCALS}. From Table~\ref{tab:pad_alg_new}, it is clear that zero padding delivers the strongest position information, compared with replicate, boundary reflection, and circular padding, supporting \textbf{H-I}. Note that partial convolution~\cite{liu2018partial} still pads with zeros, but brightening the image artificially when the convolution kernel overlaps it only partially. Thus, position information is still encoded when partial convolutions are used. Interestingly, circular padding is often the second most capable padding type. We conjecture this is because circular padding takes values from the opposite side of the image where the pixel values are typically less correlated than the directly neighbouring pixels. Thus, circular padding often has a value transition at the border, contrasting reflection and replicate which offer little or no signal to the CNN regarding the whereabouts of the image border. 

\begin{figure} [t]
	\centering
          \centering
         \resizebox{0.49\textwidth}{!}{
          \includegraphics[width=0.48\textwidth]{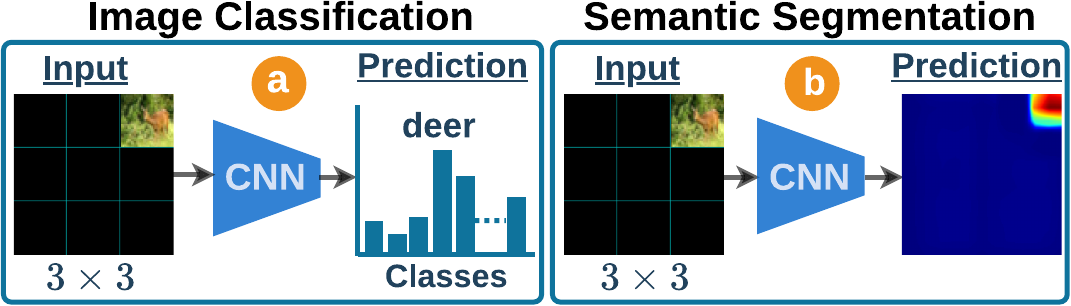} 
    }
  \caption{We consider two \textit{location dependant} tasks designed to investigate the boundary effects in CNNs. A random CIFAR-10 image is placed on a random grid location and the CNN predicts either $C$ class logits (a: classification), or $C$ class logits for each pixel (b: segmentation).}
 
  \label{fig:spectrum}
\end{figure}
\section{Location Dependant Tasks for Positional Analysis}\label{sec:loc_class_seg}
We now go deeper and explore the critical importance of the boundary effect in CNNs with respect to absolute position information by means of experiments designed to reveal these characteristics in a per-location manner.
We begin by describing our experimental settings and the implementation details for the proposed location dependant experiments with grid-based inputs. These experiments are used to analyze the border effects with respect to position information encoded in CNNs. These consist of location dependant image classification (Fig.~\ref{fig:spectrum} (a) and Sec.~\ref{sec:img_cls_with_loc}), and image segmentation (Fig.~\ref{fig:spectrum} (b) and Sec.~\ref{sec:seg_with_loc}), under different canvas color settings. Our experiments are designed with the goal of determining, for different canvas colors (\textbf{H-II}), where in the input CNNs suffer from the border effect (\textbf{H-IV}), and how the position of an image affects the learning of semantic features (\textbf{H-III}).
\begin{figure}
  \begin{center}
   \resizebox{0.47\textwidth}{!}{ 
      \includegraphics[width=0.49\textwidth]{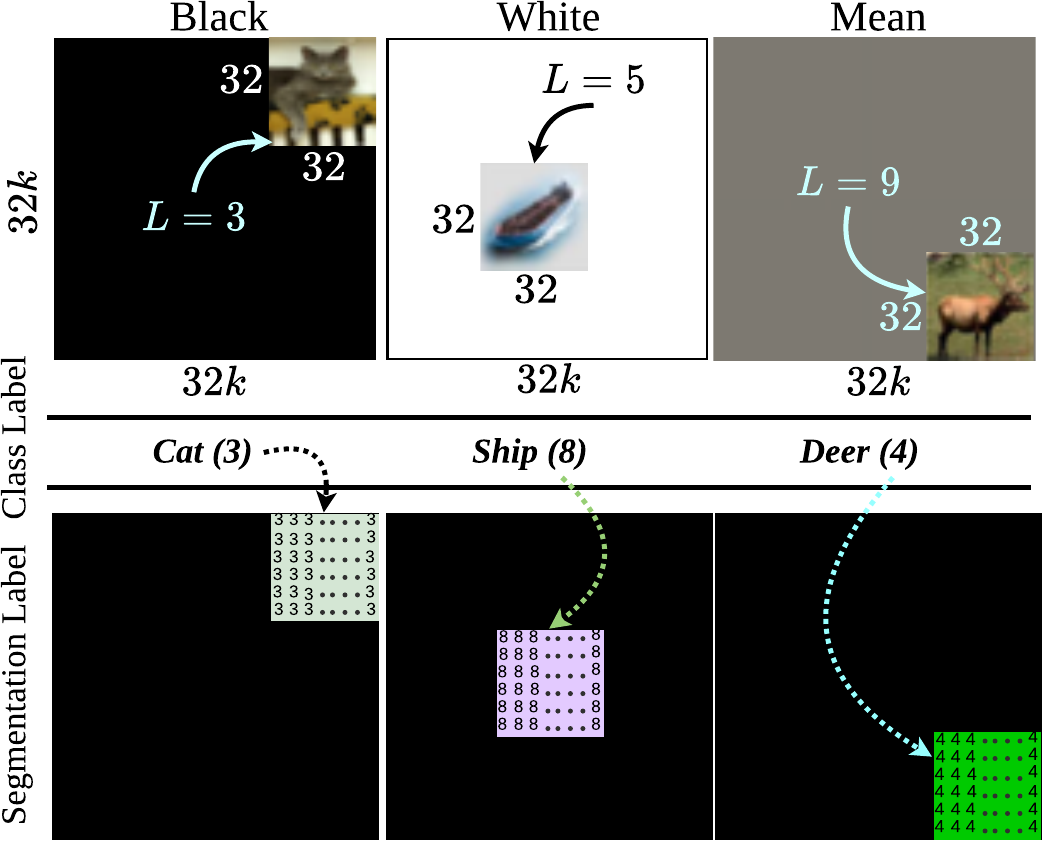} }
  \end{center}
  \vspace{-0.25cm}
  \caption{An illustration of the grid settings ($k=3$) and the ground-truth with all three canvas colors for the location dependant tasks.}
  \label{fig:grid_main}
  \vspace{-0.3cm}
\end{figure}
\subsection{Experimental Settings and Implementation Details} Our image classification and segmentation experiments use `location dependant' inputs (see Fig.~\ref{fig:grid_main}). The input is a colored canvas (the colors used are \textit{Black} $[0,0,0]$, \textit{White} $[1,1,1]$, and the CIFAR-10 dataset~\cite{krizhevsky2014cifar} \textit{Mean} $[0.491,0.482,0.446]$) with an image patch randomly placed on a $k \times k$ grid. The motivation of using different canvas colors in grid settings is inspired by ~\cite{kayhan2020translation} which paste an image patch on a black canvas to determine if a CNN can classify the image location for different resolutions (i.e., top left or bottom right). We have shown that zero padding (i.e., black) significantly increases the amount of position information encoded in the network. This suggests the border color may be playing a role in the CNNs position encoding. Thus, we paste image patches on various canvas colors and sizes with the motivation of evaluating whether the canvas color have an effect on the amount of position information encoded at various distances to the boundary. Unless mentioned otherwise, we use CIFAR-10 for all experiments. Given a $32\times32$ CIFAR-10 training image as the image patch, we \textit{randomly} choose a grid location, $L$,  and place the CIFAR-10 training sample in that location. For example, in the case of a $k\times k$ grid, the size of the grid canvas is $32k \times 32k$, where each grid location has a size of $32\times32$ and $k^2$ total locations. Figure~\ref{fig:grid_main} shows examples of inputs for the location dependant experiments, and the ground truth for each of the tasks. As previously mentioned, all the experiments were run with three different canvas colors to show the impact of the border effect with regards to canvases. Note that we normalize only the image patch before pasting it onto the canvas (in other words, the canvas does not get normalized). For the segmentation ground truth, the ratio of background pixels to object pixels grows exponentially as the grid size increases. However, as the evaluation metric is mean intersection over union (mIoU), the overall performance is averaged between the object classes and the background class, even though the background class makes up the majority of the ground truth labels. 
\begin{table*} [ht]
\caption{Location dependant (a) \text{image classification} and (b) \text{semantic segmentation} results on CIFAR-10 dataset under zero/no padding and various canvas colors (\textbf{B}lack, \textbf{W}hite, and \textbf{M}ean) settings. Note that the canvas colors have noticeable effect on image classification and segmentation performance. Additionally, the increase in performance when a black canvas is used in the no padding case compared with white or mean is particularly noteworthy.}
\centering
\def\arraystretch{1.1}
     
       \setlength\tabcolsep{4.1pt}
		\def\arraystretch{1.15}
  \resizebox{0.95\textwidth}{!}{
	\begin{tabular}{c|l|cccccc|| cccccc}
		\specialrule{1.2pt}{1pt}{1pt}\
		
		
		&\multirow{2}{*}{Padding} &  
		\multicolumn{6}{c||}{\textbf{Image Classification}} &  
		\multicolumn{6}{c}{\textbf{Image Segmentation}} \\
		\cline{3-14}
		&& \textcolor{black}{3$\times$3} & \textcolor{black}{5$\times$5} & \textcolor{black}{7$\times$7} & \textcolor{black}{9$\times$9} & \textcolor{black}{11$\times$11} & \textcolor{black}{13$\times$13} & \textcolor{black}{3$\times$3} & \textcolor{black}{5$\times$5} & \textcolor{black}{7$\times$7} & \textcolor{black}{9$\times$9} & \textcolor{black}{11$\times$11} & \textcolor{black}{13$\times$13} \\
		
		\specialrule{1.2pt}{1pt}{1pt}
		
		\multirow{2}{*}{\rotatebox[origin=c]{0}{\textbf{B}}}& 
		\text{Zero Pad}  & \textcolor{black}{\text{82.9} }& \textcolor{black}{\text{82.4}} & \textcolor{black}{\text{82.3}} & \textcolor{black}{\text{81.4}} & \textcolor{black}{\text{81.7}} & \textcolor{black}{\text{81.7}} & 
		\textcolor{black}{\text{70.9}} & \textcolor{black}{\text{68.5}} &\textcolor{black}{\text{66.7}}& \textcolor{black}{\text{65.9}}&\textcolor{black}{\text{63.1}}&\textcolor{black}{\text{62.4}}\\
		&w/o Pad  & 82.7 & 82.6 & 82.2 & 81.8 & 82.3&78.8  & 69.0&67.6&65.1&64.9&62.7&60.3\\
		
		\hline
		
		\multirow{2}{*}{\rotatebox[origin=c]{0}{\textbf{W}}}& 
		\text{Zero Pad}  & \textcolor{black}{\text{82.4}} & \textcolor{black}{\text{82.4}} & \textcolor{black}{\text{81.7}} & \textcolor{black}{\text{81.8}} & \textcolor{black}{\text{81.7}}& \textcolor{black}{\text{79.7}} &  \textcolor{black}{\text{70.4}} & \textcolor{black}{\text{68.6}} &\textcolor{black}{\text{62.9}}&\textcolor{black}{\text{61.5}}&\textcolor{black}{\text{58.8}}&\textcolor{black}{\text{52.5}}\\
		&w/o Pad  & 82.1 & 82.3 & 66.4 & 39.3 & 36.6 & 24.9& 67.5 & 63.1 &59.5&54.4&45.8&41.8\\
		\hline
		
		\multirow{2}{*}{\rotatebox[origin=c]{0}{\textbf{M}}}& 
		\text{Zero Pad}  & \textcolor{black}{\text{82.5}} & \textcolor{black}{\text{82.4}} & \textcolor{black}{\text{82.3}} & \textcolor{black}{\text{81.7}} & \textcolor{black}{\text{81.2}} & \textcolor{black}{\text{80.5}}&  \textcolor{black}{\text{70.8}}& \textcolor{black}{\text{70.8}}& \textcolor{black}{\text{65.8}}& \textcolor{black}{\text{61.7}}& \textcolor{black}{\text{62.1}}& \textcolor{black}{\text{54.8}}\\
		&w/o Pad  & 82.9 & 82.1 & 70.4 & 72.9 & 64.5 & 48.7
		& 69.2&64.0&62.7&60.3&53.7&50.0\\

		\specialrule{1.2pt}{1pt}{1pt} 
	\end{tabular}

			}

		\label{tab:semantic}

\end{table*}

\begin{table}
	\caption{Performance comparison of various no padding implementation techniques using VGG-11 network under $7 \times 7$ grid, different canvas, and task settings. 'Res` refers to the spatial resolution of the final prediction map before upsampling to the image resolution. Results show that the no padding implementation with bilinear interpolation achieves higher performance than other alternatives.} 
     \centering
    \setlength\tabcolsep{4.3pt}
	\def\arraystretch{1.15}
    \resizebox{0.49\textwidth}{!}{
         \begin{tabular}{l|c|cc|cc}
				 \specialrule{1.2pt}{1pt}{1pt}
				 \multirow{2}{*}{Padding} & \multirow{2}{*}{Res} &\multicolumn{2}{c|}{Classification} & \multicolumn{2}{c}{Segmentation} \\
				 \cline{3-6}
				   && \textbf{B} & \textbf{W} & \textbf{B} & \textbf{W}  \\
				   \specialrule{1.2pt}{1pt}{1pt}
				   Zero Pad & 7$\times$7& 84.5 &83.8 & 64.1 & 58.6\\
				   No Pad &  3$\times$3 & 80.4 & 66.5 & 9.2 & 9.6\\
				   No Pad + BI &  7$\times$7&80.6 & 70.3 & 61.9 & 49.2\\
				   
				  \specialrule{1.2pt}{1pt}{1pt}

			\end{tabular}
	}

\label{tab:nopadding}
\end{table} 
All experiments are run for $k\in\{3,5,7,9,11,13\}$. To ensure a fair comparison between grid locations, the evaluation protocol consists of running the entire validation set of CIFAR-10 on \textit{each individual} grid location (i.e., we run the validation set $k^2$ times for a single validation epoch). We then average the performance over all grid locations to obtain the overall accuracy. The motivation of using different grid sizes (smaller $\rightarrow$ larger) is to validate if absolute position can be encoded only close to image boundary or far a way from the image boundary. We report classification and segmentation accuracy in terms of precision and mean intersection over union (mIoU), respectively. We use a ResNet-18 network trained from scratch, unless stated otherwise. ResNets with no padding are achieved by setting the padding size to zero in the convolution operation. For fair comparison between the padding and no padding baseline, we use bilinear interpolation (see Sec.~\ref{sec:nopadding} for discussion) to match spatial resolutions between the residual output and the feature map for the no padding case. 
\subsection{Network Implementation Without Padding}\label{sec:nopadding}

We include no padding comparisons for completeness and to contrast the difference in the border effects between networks trained with padding and without padding. For networks without residual connections (e.g., VGG) one can implement a no padding version by simply discarding the padding. However, controlling for consistent spatial resolution is crucial when comparing padding types since an inconsistent spatial resolution between padding and no padding would result in a significant performance drop due to the reduced dimensionality of the feature representations. Another solution is to remove all the padding from a VGG network and then padding the input image by a sufficient amount to keep the spatial resolution. However, this is not applicable to the ResNet backbone as there will be spatial misalignment between the features of layers due to the residual connections. Alternatively, one can interpolate the output feature map to the same size as the input, which is also the method used in a recent study~\cite{xu2020positional}. In the end, we choose the interpolation implementation because we believe the visual information near the border is be better retained while working for networks with and without residual connections.

One concern of using interpolation is how to align the feature maps during the interpolation. If the features maps are aligned in the center, interpolating the feature map will move the contents of feature map slightly towards the edges. The composite will thus not have the features from the two branches perfectly line up with each other anymore. This shifting effect is largest near the edges and smallest near the center, which matches with the observed performance characteristics. The subsequent convolution layers may be able to undo some of this shifting, but only at the cost of location-dependent kernels that are tailored to fit the offset caused at different parts of the image. The other option is to align the feature map based on the corners with the interpolation mainly occurring at the center. In this scenario, the shifting effect will be reversed, with the corners being in alignment but the center of the feature map slightly misaligned.

To this end, we experimentally evaluate various no padding implementation techniques. We choose the VGG-11 network for this experiment since it is a lightweight network and does not contain any residual connections. Table~\ref{tab:nopadding} presents the location dependent image classification and segmentation results using VGG-11 network with $7\times7$ grid under different no padding implementation settings. Interestingly, no padding implementation with bilinear interpolation (BI) achieves superior performance than w/o BI in both the tasks; however, the performance difference is more prominent in the segmentation case as the spatial resolution of the final feature map in w/o BI case is lower than the w/ BI case which is crucial in segmentation task. Also, it seems plausible that a network could extract position information from the spatially varying slight misalignment of the feature maps (e.g., in the image center there is no misalignment and at the border there is 1 pixel of misalignment for a 3x3 convolutional layer). Taking these factors in consideration, we choose to use the bilinear interpolation-based no padding implementation in all of the following experiments.

\subsection{Location Dependant Image Classification}\label{sec:img_cls_with_loc}
We investigate whether CNNs trained with and w/o padding are equally capable of exploiting absolute position information to predict the class label in all image locations, with respect to the distance from the image boundary and for variable grid sizes. The location dependant image classification experiment is a multi-class classification problem, where each input has a single class label and the CNN is trained using the multi-class cross entropy loss (see Fig.~\ref{fig:spectrum} (a)). Therefore, the network must learn semantic features \textit{invariant to the patch location}, to reach a correct categorical assignment. 

Table~\ref{tab:semantic} (left) shows the location dependant image classification results. For all canvases, the networks trained with padding are more robust to changes in grid sizes. In contrast, models trained w/o padding under white and mean canvas settings significantly drop in performance with the increase of grid size, as position information is lost and boundary information cannot be exploited. However, when the models training w/o padding under black canvas, the classification performance results do not vary since in large grids, an image patch embedded somewhere else than the edge of a black canvas, without padding, is fundamentally the same as having just the picture and zero padding. Interestingly, the canvas colors seem to have a noticeable effect on classification performance (\textbf{H-II}). 
The difficulty in separating image semantics from the background signal is due to non-zero canvases creating noisy activations at regions near the image patch border, which is explored further in Section~\ref{sec:analysis}.
\subsection{Location Dependant Image Segmentation}\label{sec:seg_with_loc}
The experiment in this section examines similar properties as the previous location dependant image classification, but for a dense labeling scenario. This task is simply a multi-class per-pixel classification problem, where each pixel is assigned a single class label. We follow the same grid strategy as classification to generate a training sample. Since CIFAR-10 is a classification dataset and does not provide segmentation ground-truth, we generate synthetic ground-truth for each sample by assigning the class label to all the pixels in the grid location where the image belongs to (see Fig.~\ref{fig:spectrum} (b)). Following existing work~\cite{chen2017rethinking}, we use a per-pixel cross entropy loss to train the network and upsample the prediction map to the target resolution using bilinear interpolation. For evaluation, we compute mIoU at per grid location and take the average to report results.

Image segmentation results are shown in Table~\ref{tab:semantic} (right). A similar pattern is seen as the classification experiment (Sec.~\ref{sec:img_cls_with_loc}). Networks trained with padding consistently outperform networks trained w/o padding, and the difference grows larger as the grid size increases. Contrasting the classification experiment, the performance of networks with padding decreases slightly as the grid size increases. The reason for this is that the mIoU metric is averaged across all categories including the background, so object pixels are equally weighted in the mIoU calculation even though the ratio of background pixels to object pixels increases dramatically for larger grid sizes. For the no padding case, we observe similar patterns to the classification experiment as the white and mean canvas scenarios suffer more from a large grid size than the black canvas case. This finding further suggests that, independent of the task, a black canvas injects more location information to a CNN (\textbf{H-II}), regardless of the semantic difficulty, than a white or mean colored canvas, which is further explored in Sec.~\ref{sec:analysis}. 

\begin{table}
\caption{Performance comparison between ResNet18 and BagNet variants to demonstrate the relationship between these networks in terms of encoding position information. Interestingly, BagNets can classify images in absolute locations furthest away from the boundary but fail to precisely segment objects far from the boundary.}
     \centering
     
    \setlength\tabcolsep{4.3pt}
	\def\arraystretch{1.25}
    \resizebox{0.47\textwidth}{!}{
         \begin{tabular}{l|cc|cc}
				 \specialrule{1.2pt}{1pt}{1pt}
				 \multirow{2}{*}{Network} & \multicolumn{2}{c|}{Image Classification} & \multicolumn{2}{c}{Segmentation} \\
				 \cline{2-5}
				   & \textbf{B} & \textbf{W} & \textbf{B} & \textbf{W} \\
				   \specialrule{1.2pt}{1pt}{1pt}
				   ResNet18~\cite{he2016deep}& 82.4 & 82.4 & 68.5 & 68.6\\
				   
				   ResNet50~\cite{he2016deep}& 83.1 & 83.2 & 70.1 & 69.7\\
				   \hline 
				   BagNet33~\cite{brendel2019approximating} & 82.7& 81.4 & 30.4 & 32.2\\
				   BagNet17~\cite{brendel2019approximating} & 80.6 & 80.7& 34.5 & 34.7 \\
				   
				    BagNet9~\cite{brendel2019approximating} & 70.1 &66.8 &30.6 & 28.7  \\
				  \specialrule{1.2pt}{1pt}{1pt}

			\end{tabular}
	}
	
\label{tab:bagnet}
\end{table} 
\begin{figure}
	\begin{center}
		\includegraphics[width=0.49\textwidth]{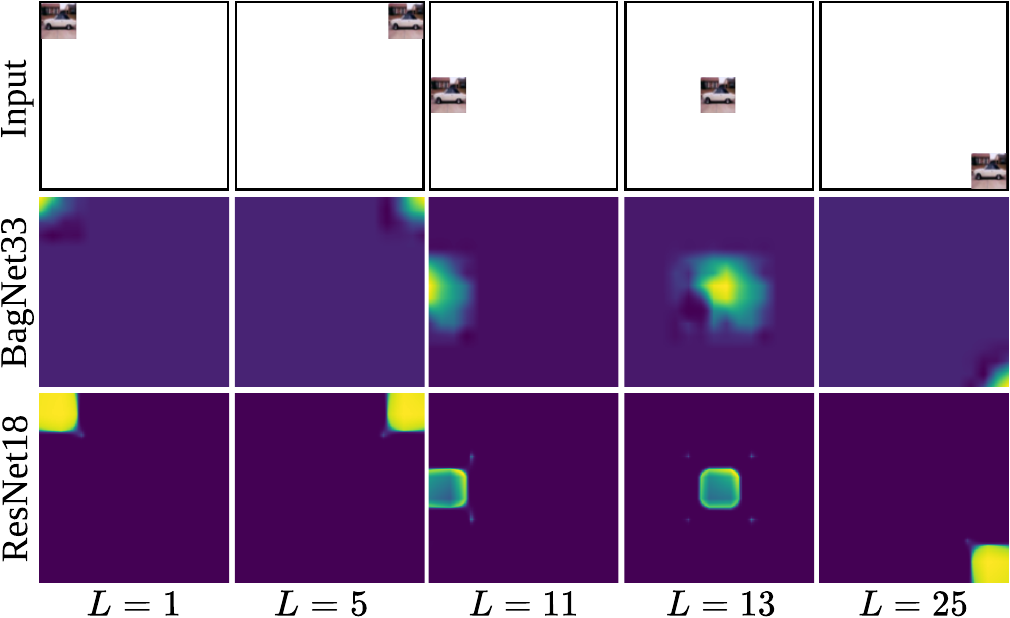}
		\caption{Comparison of BagNet33 and ResNet18  semantic segmentation results on different locations of a $5\times5$ grid under the white canvas setting. Confidence maps are plotted with the ‘cividis’ colormap, where yellow and blue indicates higher and lower confidence, respectively.}
				
		\label{fig:bagnet_pred}
	\end{center}
	\vspace{-0.5cm}
\end{figure}
\subsection{Relationship Between Receptive Field Size and Boundary Effects}
Our prior work~\cite{islam2020much} studies the impact of varying kernel sizes in the position encoding readout module while extracting absolute position information from a pretrained CNN. The results suggest that larger kernel sizes are likely to capture more position information than smaller sizes. A logical next line of inquiry from these results is how the \textit{receptive field} of a network effects the ability to encode position information. To this end, we now evaluate the relationship between a network's effective receptive field and its ability to encode position information by comparing two types of networks, ResNets and BagNets~\cite{brendel2019approximating}). BagNets are a modified version of ResNet50 that restrict the effective receptive field of the CNN to be a fixed maximum, i.e., either 9, 17, or 33 pixels. 
\begin{figure} [t]
	\centering
          \centering
         \resizebox{0.45\textwidth}{!}{
          \includegraphics[width=0.45\textwidth]{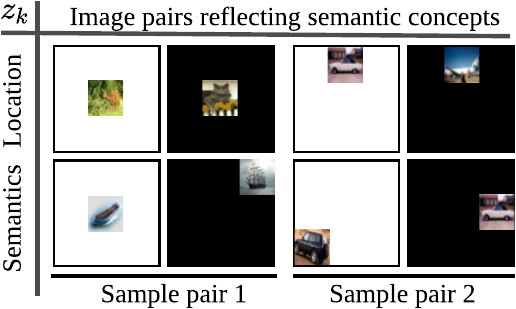} 
    }
  \vspace{-0.25cm}
  \caption{Sample pair generation reflecting two semantic concepts (location and semantic class).}
  \label{fig:dim_sample}
\end{figure} 
\begin{table} [t]
\caption{Dimensionality estimation (\%) of two semantic concepts (location and semantic category) under different tasks and settings. Networks trained with zero-padding and black canvas encode more location specific dimensions compared to white canvas and no padding.}
	\def\arraystretch{1.25}
	\setlength\tabcolsep{3.2pt}
		\resizebox{0.49\textwidth}{!}{
			\begin{tabular}{c |l|l|cc|cc}
				\specialrule{1.2pt}{1pt}{1pt}
				\multirow{2}{*}{\hspace{0.01cm} \rotatebox{0}{$\ast$}}&
				\multirow{2}{*}{\hspace{0.01cm} Grid}&
				\multirow{2}{*}{\hspace{0.01cm} Padding}& \multicolumn{2}{c|}{Segmentation} & 
				\multicolumn{2}{c}{Classification} \\
				\cline{4-7}
				
				&&& $|z_{\text{Location}}|$ & $|z_{\text{Class}}|$ & $|z_{\text{Location}}|$ & $|z_{\text{Class}}|$ \\
				\specialrule{1.2pt}{1pt}{1pt}	
				
				 \multirow{2}{*}{\textbf{B}} &\multirow{2}{*}{ 7$\times$7} & Zero Pad & \textbf{15.2\%} & \textbf{14.9\%} & \textbf{12.7\%} & \textbf{12.6\%}\\
				
				&& No Pad & 12.7\% & 12.8\% & 12.1\% & 11.9\%\\
				\hline
				
				
				\multirow{2}{*}{\textbf{W}} &\multirow{2}{*}{ 7$\times$7} & Zero Pad & \textbf{12.5\%} & \textbf{12.3\%} & \textbf{12.2\%} & \textbf{12.1\%}\\
				
				&& No Pad& 10.9\% & 10.9\% &11.5\% & 11.6\% \\

				\specialrule{1.2pt}{1pt}{1pt}
			\end{tabular}
		
			}

		\label{tab:dim}
\end{table}
The results of this comparison are presented in Table~\ref{tab:bagnet} where both the ResNet50 and the BagNet variants are trained on CIFAR-10 for location dependent image classification and segmentation under different canvas settings. Interestingly, BagNets variants can classify image positioned further away from the boundary similar to the ResNet18 network. Note that the image patch size is $32\times32$ and so the receptive field of the BagNet33 and 17 can cover a large portion of the patch. This is why the BagNet9 suffers more in performance. For semantic segmentation, the performance is significantly lower for all BagNet variants. These results show that the network with larger receptive field and zero padding can handle boundary effects more effectively by exploiting absolute position information.

Figure~\ref{fig:bagnet_pred} shows the probability heatmaps of BagNet33 and ResNet18 segmentation predictions for different grid locations, $L$, for a $5\times5$ grid. Due to the restricted receptive field, BagNet33 have difficulty segmenting images precisely particularly near the border. In summary, there is a strong correlation between boundary effects and effective receptive field size in the absolute position encoding in CNNs. 
\section{Interpreting Representations for  Dimensionality Estimation}\label{sec:dim_est}
Previous works~\cite{bau2017network,esser2020disentangling,islam2020shape} proposed various mechanisms to interpret different semantic concepts from latent representations by means of quantifying the number of neurons which encode a particular semantic factor, $k$. Given a pretrained CNN encoder $\E(I)=z$ where $z$ is a latent representation and given an image pair ($I^a$, $I^b$) $\sim$ $p(I^a$, $I^b | k)$ which are similar in the $k$-th semantic concept, we aim to estimate the dimensionality of the semantic factor, $z_k$, that represents this concept in the latent representation. A positive mutual information between $I^a$ and $I^b$ implies a similarity of $I^a$ and $I^b$ in the $k$-th semantic concept, which will be preserved in the latent representations $\E(I^a)$ and $\E(I^b)$, only if $\E$ encodes the $k$-th semantic concept. Following~\cite{esser2020disentangling}, we approximate the mutual information between $\E(I^a)$ and $\E(I^b)$ with the correlation of each dimension, $i$, in the latent representation.

\begin{equation}
    \text{Correlation}_k=C_k =\sum_{i} \frac{\operatorname{Cov}\left(\E\left(I^a\right)_{i}, \E\left(I^b\right)_{i}\right)}{\sqrt{\operatorname{Var}\left(\E\left(I^a\right)_{i}\right) \operatorname{Var}\left(\E\left(I^b\right)_{i}\right)}},
\end{equation}

We assume that the residual factor has a maximum dimension of $|z|$ (the total dimension of the latent representation) and use the softmax equation to get the resulting dimension: 

\begin{equation}
    |z_{k}|=\left\lfloor\frac{\exp C_k}{\sum_{f=0}^{F} \exp C_f} N\right\rfloor ,
\end{equation}

\noindent where $|z_{k}|$ is the dimension of the semantic factor $k$, and $F$ is the total number of semantic factors including the residual factor. Note we do not need an estimate of the absolute mutual information for estimating the proportion of location and semantic dimensions. Only the differences between the mutual information for position and semantic class for image pairs are used to quantify the ratio of location and semantic-specific neurons. Therefore, the relative difference is still meaningful and only the absolute numbers might not be. 

We generate image pairs which share one of two semantic concepts: (i) \textit{location} or (ii) \textit{semantic class}. For example, the image pair sharing the location factor (see Fig.~\ref{fig:dim_sample} top row) differs in the class and canvas color, while the pair on the bottom row shares the semantic class but differs in canvas color and location. With this simple generation strategy, we can accurately estimate the number of dimensions in the latent representation which encodes the $k$-th semantic factor. Note that the remaining dimensions not captured in either the location or semantic class is allocated to the \textit{residual} semantic factor, which by definition will capture all other variability in the latent representation, $z$. 

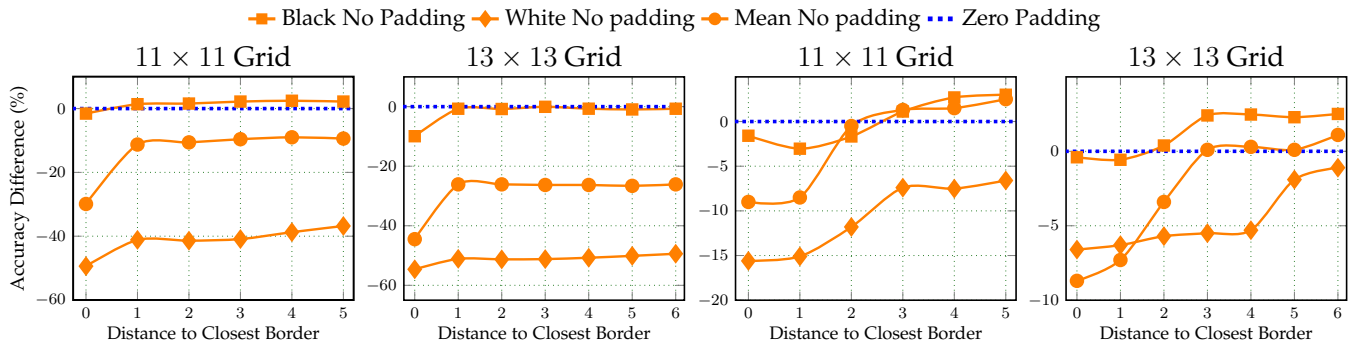
\begin{figure*}
	\begin{center}
	\begin{minipage}{0.99\linewidth}\ref{named_dist_diff}
     \centering 
		\resizebox{1.0\textwidth}{!}{ \vspace{-0.7cm}
\begin{tikzpicture} 
    \begin{axis}[
       line width=1.0,
        title={$11\times 11$ Grid},
        title style={at={(axis description cs:0.5,1.12)},anchor=north,font=\Large},
        xlabel={Distance to Closest Border},
        ylabel={Accuracy Difference (\%)},
        xmin=-0.25, xmax=5.2,
        ymin=-60, ymax=10,
        xtick={0,1,2,3,4,5,6},
        ytick={0,-20,-40,-60},
        x tick label style={font=\footnotesize},
        y tick label style={font=\footnotesize},
        x label style={at={(axis description cs:0.5,0.05)},anchor=north,font=\small},
        y label style={at={(axis description cs:0.12,.5)},anchor=south,font=\normalsize},
        width=6.5cm,
        height=5.5cm,        
        ymajorgrids=true,
        xmajorgrids=true,
        major grid style={dotted,green!40!black},
        legend style={
         draw=none,
         nodes={scale=0.90, transform shape},
         cells={anchor=west},
         legend style={at={(-14.3,8.9)},anchor=north}},
         legend image post style={scale=0.6},
         legend columns=-1,
         legend entries={[black]Black No Padding,[black]White No padding,[black]Mean No padding, [black]Zero Padding},
        legend to name=named_dist_diff,
    ]
    \addplot[line width=1.2pt,mark size=2.5pt,smooth,color=orange,mark=square*,]
        coordinates {(0,-1.57)(1,1.375)(2,1.571)(3,2.2)(4,2.45)(5,2.2)};
    \addplot[line width=1.2pt,mark size=4pt,smooth,color=orange,mark=diamond*,]
        coordinates {(0,-49.39)(1,-41.15)(2,-41.41)(3,-40.9)(4,-38.74)(5,-36.8)};
    \addplot[line width=1.2pt,mark size=3pt,smooth,color=orange,mark=*,]
        coordinates {(0,-29.9)(1,-11.275)(2,-10.6)(3,-9.6)(4,-9)(5,-9.4)};
    \addplot[line width=1.8pt, blue,dotted, sharp plot,update limits=false] 
	    coordinates {(-5,0)(10,0)};
        
    \end{axis}
\end{tikzpicture}

\begin{tikzpicture}
    \begin{axis}[
      line width=0.8,
        title={$13\times 13$ Grid},
        title style={at={(axis description cs:0.5,1.12)},anchor=north,font=\Large},
        xlabel={Distance to Closest Border},
        xmin=-0.25, xmax=6.2,
        ymin=-65, ymax=10,
        xtick={0,1,2,3,4,5,6},
        ytick={0,-20,-40,-60},
        x tick label style={font=\footnotesize},
        y tick label style={font=\footnotesize},
        x label style={at={(axis description cs:0.5,0.05)},anchor=north,font=\small},
        ylabel near ticks,
        width=6.5cm,
        height=5.5cm,
        ymajorgrids=true,
        xmajorgrids=true,
        major grid style={dotted,green!40!black},
    ]
    \addplot[line width=1.2pt,mark size=2.5pt,smooth,color=orange,mark=square*,]
        coordinates {(0,-9.93)(1,-.6975)(2,-.79125)(3,-.0542)(4,-.6875)(5,-0.9375)(6,-0.7)};
    \addplot[line width=1.2pt,mark size=4pt,smooth,color=orange,mark=diamond*,]
        coordinates {(0,-54.62)(1,-51.1)(2,-51.28)(3,-51.2)(4,-50.74)(5,-50.11)(6,-49.4)};
    \addplot[line width=1.2pt,mark size=3pt,smooth,color=orange,mark=*,]
        coordinates {(0,-44.5)(1,-26.1)(2,-26.1)(3,-26.3)(4,-26.3)(5,-26.6)(6,-26.1)};
    \addplot[line width=1.8pt, blue,dotted,sharp plot,update limits=false] 
	    coordinates {(-5,0)(10,0)};
        \end{axis}
\end{tikzpicture}

\begin{tikzpicture}
    \begin{axis}[
       line width=0.8,
        title={$11\times 11$ Grid},
        title style={at={(axis description cs:0.5,1.12)},anchor=north,font=\Large},
        xlabel={Distance to Closest Border},
        xmin=-0.25, xmax=5.2,
        ymin=-20, ymax=5,
        xtick={0,1,2,3,4,5},
        ytick={0,-5, -10,-15, -20},
        x tick label style={font=\footnotesize},
        y tick label style={font=\footnotesize},
        x label style={at={(axis description cs:0.5,0.05)},anchor=north,font=\small},
        ylabel near ticks,
        width=6.5cm,
        height=5.5cm,        
        ymajorgrids=true,
        xmajorgrids=true,
        major grid style={dotted,green!40!black},
    ]
    \addplot[line width=1.2pt,mark size=2.5pt,smooth,color=orange,mark=square*,]
        coordinates {(0,-1.59)(1,-3.04)(2,-1.67)(3,1.12) (4,2.6975)(5,3)};
    \addplot[line width=1.2pt,mark size=4pt,smooth,color=orange,mark=diamond*,]
        coordinates {(0,-15.6)(1,-15.1)(2,-11.8)(3,-7.4) (4,-7.51)(5,-6.6)};
    \addplot[line width=1.2pt,mark size=3pt,smooth,color=orange,mark=*,]
        coordinates {(0,-9.0)(1,-8.5)(2,-0.5)(3,1.3) (4,1.5)(5,2.5)};
    \addplot[line width=1.8pt, blue,dotted,sharp plot,update limits=false] 
	    coordinates {(-5,0)(10,0)};
    \end{axis}
\end{tikzpicture}

\begin{tikzpicture}
    \begin{axis}[
      line width=0.8,
        title={$13\times 13$ Grid},
        title style={at={(axis description cs:0.5,1.12)},anchor=north,font=\Large},
        xlabel={Distance to Closest Border},
        xmin=-0.25, xmax=6.2,
        ymin=-10, ymax=5,
        xtick={0,1,2,3,4,5,6},
        ytick={0,-5,-10},
        x tick label style={font=\footnotesize},
        y tick label style={font=\footnotesize},
        x label style={at={(axis description cs:0.5,0.05)},anchor=north,font=\small},
        ylabel near ticks,
        width=6.5cm,
        height=5.5cm,
        ymajorgrids=true,
        xmajorgrids=true,
        major grid style={dotted,green!40!black},
    ]
    \addplot[line width=1.2pt,mark size=2.5pt,smooth,color=orange,mark=square*,]
        coordinates {(0,-0.4)(1,-0.567)(2,0.39)(3,2.41) (4,2.47) (5,2.29) (6,2.51)};
    \addplot[line width=1.2pt,mark size=4pt,smooth,color=orange,mark=diamond*]
        coordinates {(0,-6.6)(1,-6.3)(2,-5.7)(3,-5.5)(4,-5.3)(5,-1.9)(6,-1.1)};
    \addplot[line width=1.2pt,mark size=3pt,smooth,color=orange,mark=*,]
        coordinates {(0,-8.7)(1,-7.3)(2,-3.4)(3,0.1)(4,0.3)(5,0.1)(6,1.1)};
    \addplot[line width=1.8pt, blue,dotted,sharp plot,update limits=false] 
	    coordinates {(-5,0)(10,0)};
        \end{axis}
    \end{tikzpicture}
   }
   \end{minipage}
      
    	
	\end{center}
	\vspace{-0.3cm}
	\caption{Location dependant image classification (left two) and segmentation (right two). Results show the accuracy difference between padding and no padding under three canvas settings, at various distances to the border.}
	\label{fig:border}
	\vspace{-0.1cm}
\end{figure*}
        
Table~\ref{tab:dim} shows the estimated dimensionality for the semantic factors \textit{location} and \textit{class}. The latent representation used is the last stage output of a ResNet-18 before the global average pooling layer. We used the networks from Sec.~\ref{sec:loc_class_seg} which are trained for segmentation (left) and classification (right) with the appropriate background (i.e., black on the top and white on the bottom row) and grid settings. The results clearly show that networks trained with zero-padding contain more dimensions which encode the semantic factor `location' (\textbf{H-I}). Further, Table~\ref{tab:dim} shows that there is a \textit{positive correlation} between the encoding of location and the encoding of semantics, i.e., a larger number of dimensions encoding location implies a larger number of neurons encoding semantics, supporting \textbf{H-III}. 

\section{Per-Location Analysis}\label{sec:analysis}
In this section, we take advantage of the grid-based learning paradigm and conduct further evaluations on a per-location basis to test \textbf{H-I}, \textbf{H-II}, \textbf{H-III}, and \textbf{H-IV}. In particular, we analyze the \textit{relationship} between zero padding and the border effect. We then show quantitative and qualitative results which reveal strong evidence that \textit{zeros}, whether as a canvas or padding, inject maximal location bias. 
\subsection{Distance-to-Border Analysis: What Input Regions Suffer Most from Border Effects?} First, we analyze the image classification and segmentation results reported in Secs.~\ref{sec:img_cls_with_loc} and~\ref{sec:seg_with_loc}, with respect to the distance from the closest border which will allow us to answer this question. To obtain the accuracy at each distance, we average the accuracies over all grid locations with the same distance to the nearest border (e.g., a distance to a border of zero refers to the average accuracy of the outer-most ring of grid locations). Figure~\ref{fig:border} shows the accuracy difference between the padding baseline (the blue horizontal line) and the no padding cases. Interestingly, the accuracy difference is higher at grid locations close to the border and decreases towards the image center. This analysis strongly suggests that zero padding significantly impacts the border effect, and injects position information to the network as a function of the object location relative to the distance of the nearest border. In contrast, the no padding case fails to deliver any position information at the border locations which leads to a significant performance drop. Also note that there is a substantial difference in performance at the center of the image, at the farthest distance from the border, supporting \textbf{H-IV}. 
Note that of the three canvases for the no padding case, the black canvas yields the lowest drop in relative performance when comparing the center region to locations near the border (\textbf{H-II}). More distance-to-border analysis results can be found in Sec. A.3.2 in the appendix.
\begin{figure}
	\begin{center}
		\includegraphics[width=0.48\textwidth]{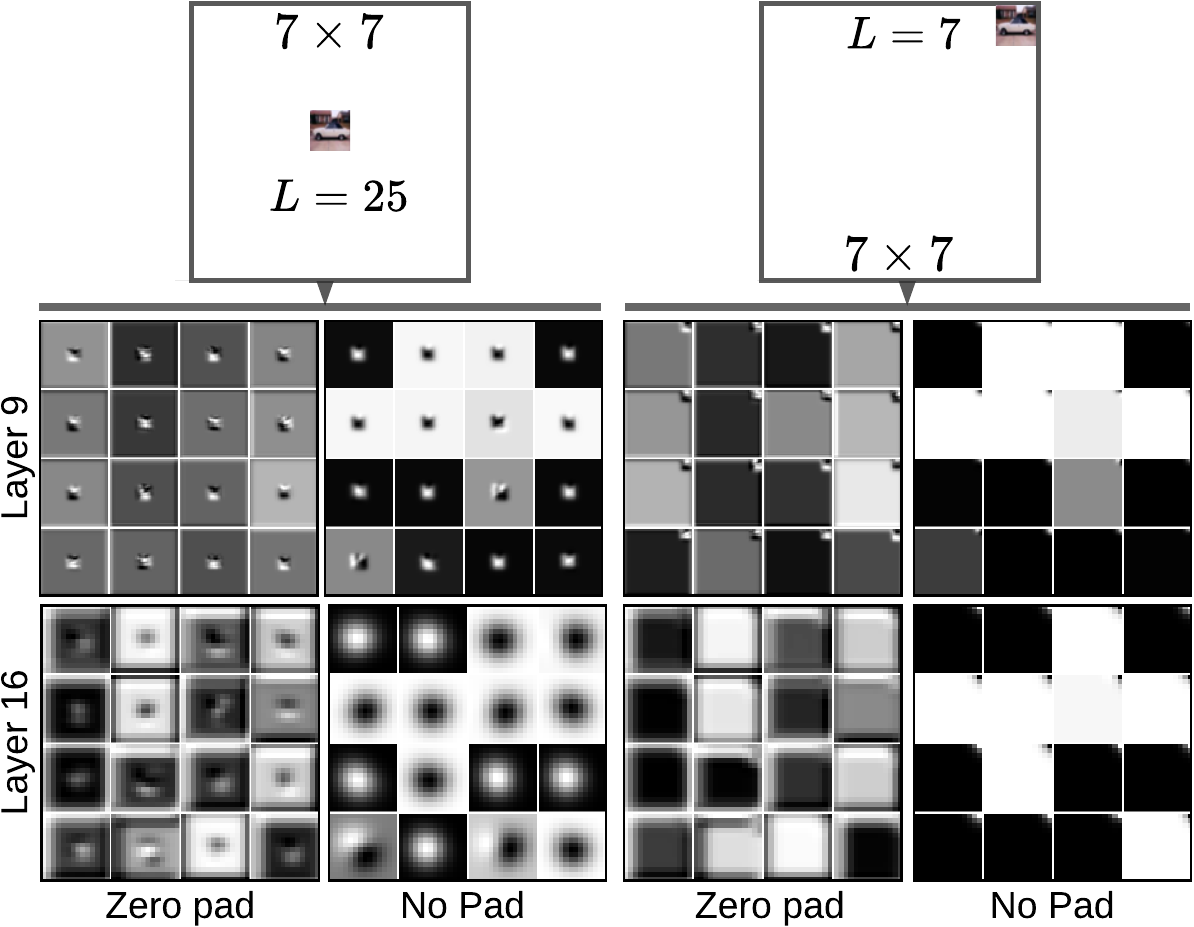}
		\vspace{-0.1cm}
		\caption{Filter activation visualization for the classification task on CIFAR-10 with a \textit{white} background and $7\times7$ grid size. It is clear that zero padding provides richer information and larger activations downstream, particularly at locations near the boundary (e.g., $L=7$). The activations are visualized using the `gray' colormap.}
				
		\label{fig:border_activation}
	\end{center}
	\vspace{-0.5cm}
\end{figure}
\begin{figure*}
	\begin{center}
		\includegraphics[width=0.99\textwidth]{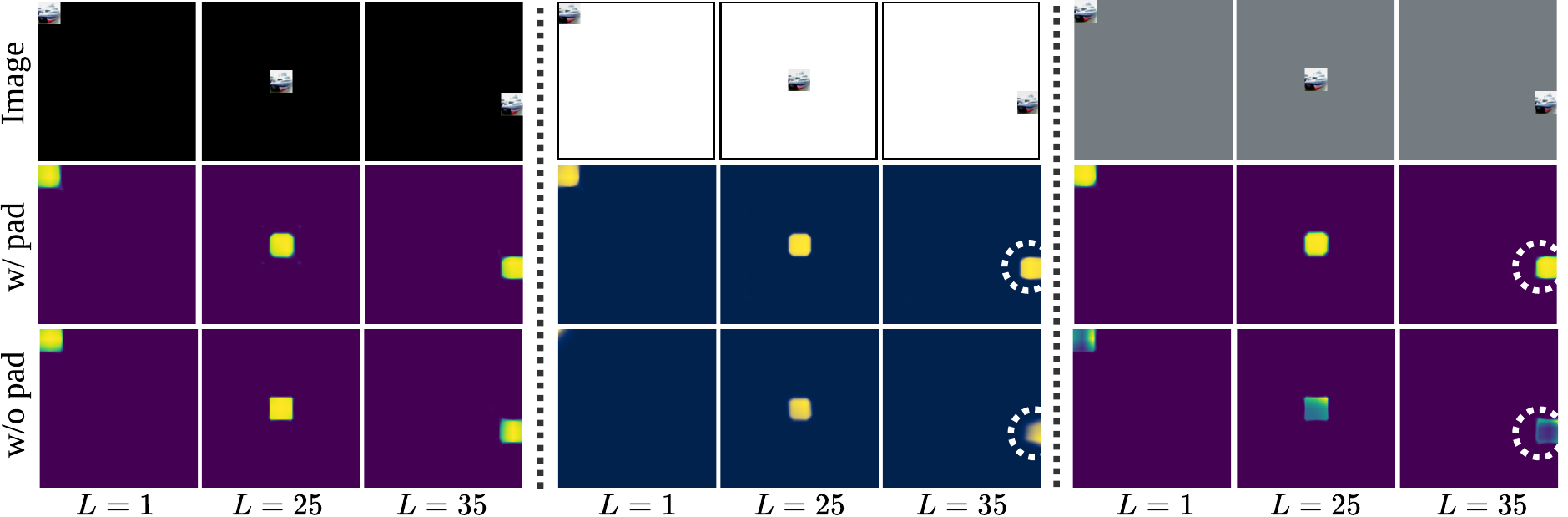}
		\caption{Sample predictions of semantic segmentation on different locations of a $7\times7$ grid under three background settings. Confidence maps are plotted with the ‘cividis’ colormap, where yellow and dark blue indicates higher and lower confidence, respectively. Clearly, the encoding of absolute position information, by means of zero padding or a black canvas, has a stark effect on a CNN's ability to segment semantic regions by learning distinctive features.}
				
		\label{fig:seg_pred}
	\end{center}
	\vspace{-0.3cm}
\end{figure*}
\subsection{Are Border Effects Only at the Border?} While intuition might suggest the border effect occurs solely at the border, it is natural to analyze if other regions in the input space also suffer from the border effect. Figure~\ref{fig:border_activation} compares filter activations with and without zero padding. Note that filter activations are randomly sampled from the feature map for the specific layer. Activations found near the border propagate less information through the network during the forward pass due to the limited amount of connectivity to downstream layers compared to activations at the center, as discussed in~\cite{tsotsos1995modeling}. Further, the convolution cannot fully overlap the border regions without padding and thus will not recognize objects as well. This phenomenon can be seen in Fig.~\ref{fig:border_activation} (bottom-right), where the activations for grid location 7 are significantly reduced in the no padding case. Interestingly, for grid location 25 (i.e., center), there is also a visible difference in the activation space. Here, activations found for the no padding case are blurred and noisy which contrasts the tight square shaped activations when zero padding is used. While border effects mainly impact regions near the border, these results show clear evidence that input locations at the center of the image are also impacted with a lack of padding which is evidence supporting \textbf{H-IV}. This also explains the performance drop at the center of the grid in Fig.~\ref{fig:border} (left).


\subsection{Does Encoding Location Enable the Learning of Semantics?} In Sec.~\ref{sec:dim_est}, we provided quantitative evidence that reveals the correlation between the number of neurons encoding position and semantic information \textbf{(H-III)}. We further investigate this phenomenon to see how position information, by means of zero padding, allows for richer semantics to be learned for the tasks of image classification and semantic segmentation. 
The heatmaps in Fig.~\ref{fig:seg_pred} show segmentation predictions for different grid locations, $L$, of a $7\times7$ grid. 
When no padding is used CNNs have difficulty segmenting images near the border (highlighted with circles in Fig.~\ref{fig:seg_pred}) except when a black canvas is used. However, for locations near the center of the image, reduced position information due to no padding greatly reduces the network's confidence in semantic encodings. In contrast, zero padding is consistent and confident in segmenting objects across all the grid locations and canvas colors. 
\begin{figure}
\centering
  \begin{center}
      \includegraphics[width=0.49\textwidth]{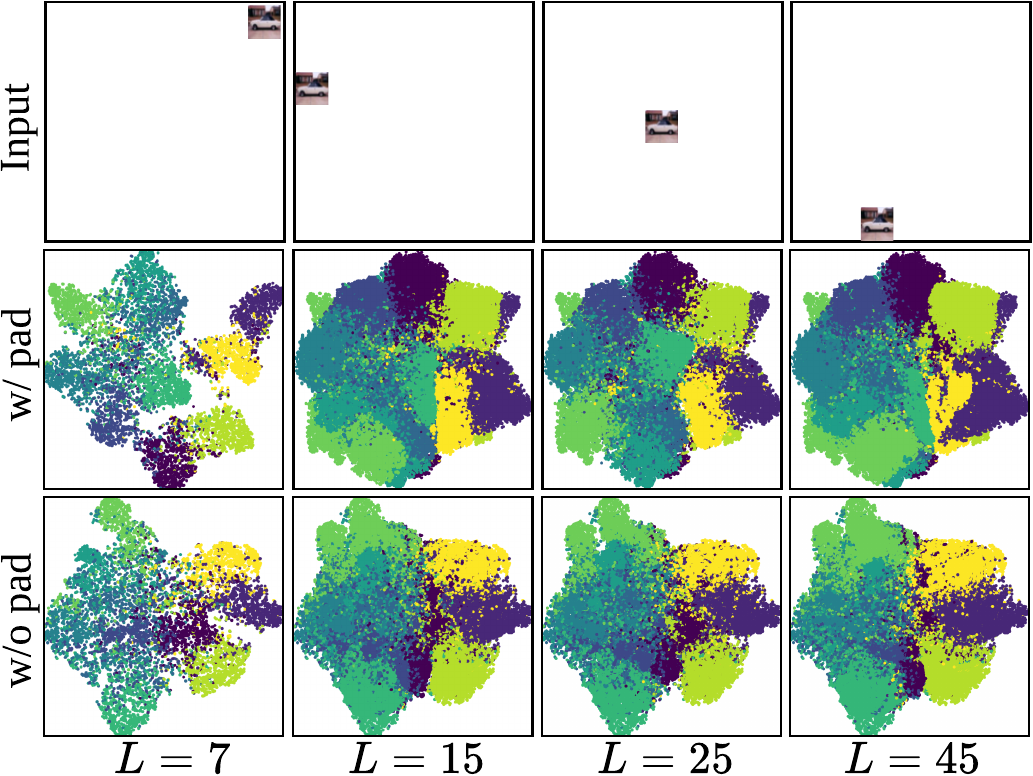} 
  \end{center}
  \vspace{-0.2cm}
  \caption{t-SNE~\cite{maaten2008visualizing} visualization of the CIFAR-10 test set classification logits for a $7\times7$ grid. Examples of a single input are given in the top row, while the embedding visualizes the entire dataset (bottom two rows). The semantic separability effect is particularly pronounced at location $L=7$.}
  \label{fig:tsne_new}
  \vspace{-0.3cm}
\end{figure}
Further, we use t-SNE~\cite{maaten2008visualizing} to visualize the classification logits in Fig.~\ref{fig:tsne_new}. 
Note that the single input examples at the top row are shown merely to highlight the location $L$, and that the second and third rows show embeddings of the entire test set. The separability of the semantic classes is significantly improved when padding is used, and the effect is particularly pronounced at locations near the border ($L=7$). This further supports the hypothesis that absolute position information, by means of zero padding, enables CNNs to learn more robust semantic features, which in turn allows for greater separability in the prediction logits. More analysis results can be found in Sec. A.4 in the appendix.


\subsection{Canvas Analysis: Why Do Explicit Zeros Inject Location Information?}
\begin{figure} [t]
	\begin{center}
		\includegraphics[width=0.47\textwidth]{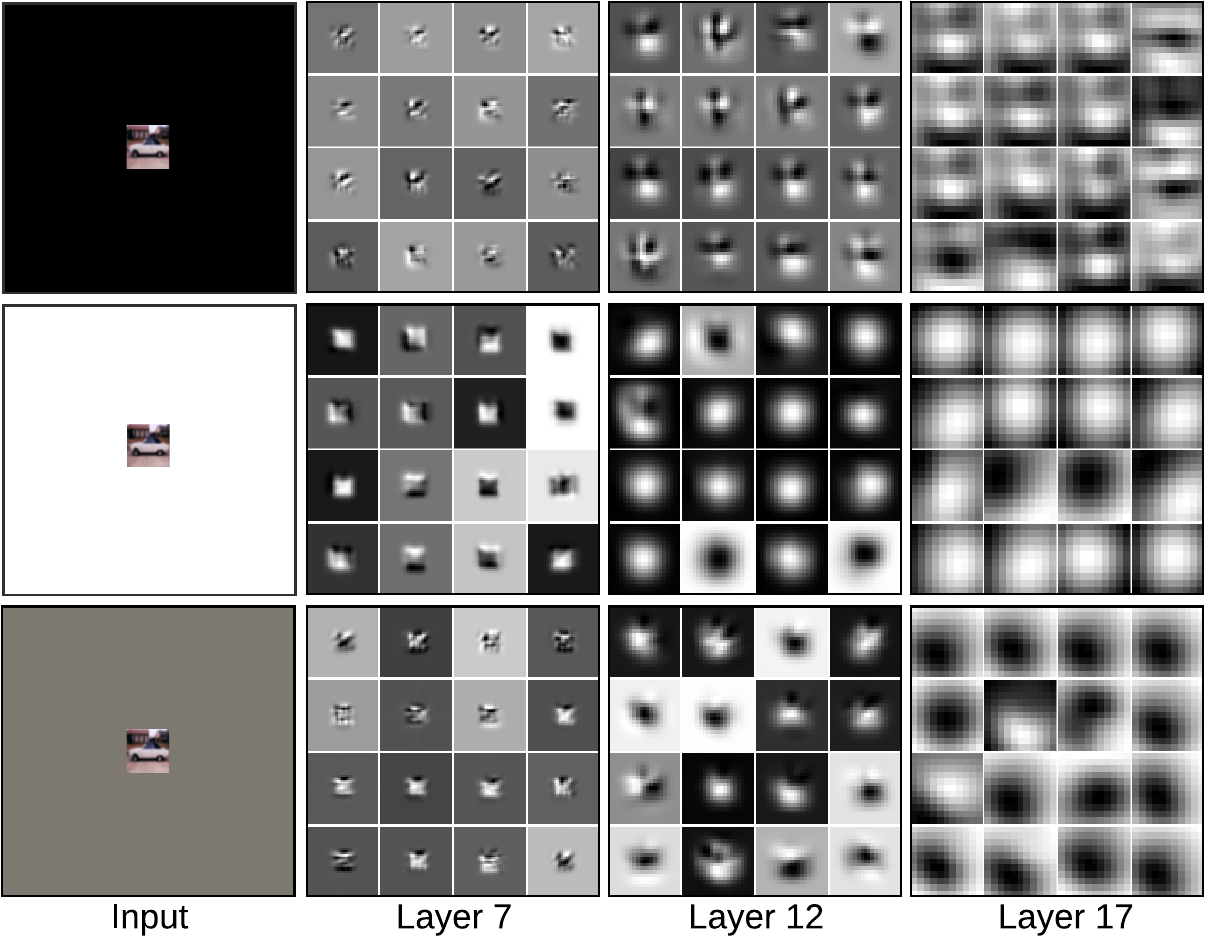}
		\vspace{-0.20cm}
		\caption{Comparison of \text{filter activations} (visualized using `gray' colormap) for the location dependant segmentation task trained without padding, $5\times5$ grid size, $L=13$, and three canvas colors, black, white, and mean. Notice the large activations in the background region for black, contrasting that of white and mean.}
		\label{fig:background_diff}
	\end{center}
	\vspace{-0.5cm}
\end{figure}

We now explore what enables CNNs to encode positional information when zeros exist at the boundary (i.e., as padding or canvas (\textbf{H-I})) by analyzing the activations of a network trained for the location dependant segmentation task. For a $k\times k$ grid, the ratio of canvas pixels to total pixels is $\frac{k^2-1}{k^2}$. This implies that the vast majority of labels will be the background class, and therefore the majority of filters should focus on correctly labelling the canvas. To determine if this is true for all canvases, we visualize randomly sampled filter activations (see Fig.~\ref{fig:background_diff}) for networks trained \textit{without} padding for the location dependant segmentation task. The activations are visualized using the `gray' colormap, where light and dark intensities denote high and low activations, respectively. Note that the activations are taken from the output of the convolutional layer and are normalized to between $[0,1]$ before plotting. Even at the earlier layers (e.g., layer 7), there is a clear difference in the patterns of activations. The majority of filters have low activations for the image region, but \textit{high activations for the background region}. In contrast, the \textit{white and mean canvases have mostly low activations for the canvas} but high activations for the image. Interestingly, particularly at layer 17 (the last convolution layer), the activations for the black background are reminiscent of oriented filters (e.g., Gaussian derivative filters) in a number of \textit{different orientations and locations}, indicating they can capture more diverse input signals compared to the white and mean canvases, which consistently activate over the \textit{center} of the input region. Figure~\ref{fig:background_diff} clearly demonstrates that zeros at the boundary, in the form of a black canvas, allows easier learning of semantics and absolute position for CNNs compared to other values supporting \textbf{H-II}.

In summary, we have shown strong evidence that despite the image boundary suffering the most, \textit{all} regions in the input are impacted by the \textit{boundary effect} with a lack of zero padding (\textbf{H-IV}). Further, using zero padding to combat border effects and encode position information concurrently enables CNNs to also learn richer and more separable \textit{semantic} features (\textbf{H-III}). Finally, we visualized features for different canvases, and showed that explicit zeros (in the form of a black canvas), allows for easier learning of semantic and location information in CNNs (\textbf{H-II}).
\begin{table} [ht]
\caption{Comparison of mIoU with DeepLabv3 using various padding types for different image regions. Top-left image in Fig.~\ref{fig:city} shows outer regions used for this analysis. The performance drop at the border region is more pronounced for no or reflect padding case than zero padding.}
\centering
\def\arraystretch{1.25}
     \centering
    
       \setlength\tabcolsep{4.1pt}
		\def\arraystretch{1.2}
    \resizebox{0.46\textwidth}{!}{
         \begin{tabular}{l|cccc}
				 \specialrule{1.2pt}{1pt}{1pt}
				  \multirow{2}{*}{Padding} &   \multicolumn{4}{c}{Evaluation Region mIoU(\%)}\\
				  \cline{2-5}
				  & 0\% - 5\%& 5\% - 10\% & 10\% - 15\% & 100\% \\
				  \specialrule{1.2pt}{1pt}{1pt}
				  
				  \textbf{Zero Pad}  & \textbf{72.6} & \textbf{72.7}& \textbf{73.8}& \textbf{74.0} \\
				  Reflect Pad& 71.9& 72.0 & 73.7 & 73.9\\
				  No Pad  & 63.7 & 66.4&67.3 & 69.1\\
				\specialrule{1.2pt}{1pt}{1pt}
			\end{tabular}
	}
	 
\vspace{-0.2cm}
\label{tab:seg_city}
\end{table}
\begin{figure*}
     \centering
	\begin{center}

	\resizebox{0.94\textwidth}{!}{
          \includegraphics[width=0.94\textwidth]{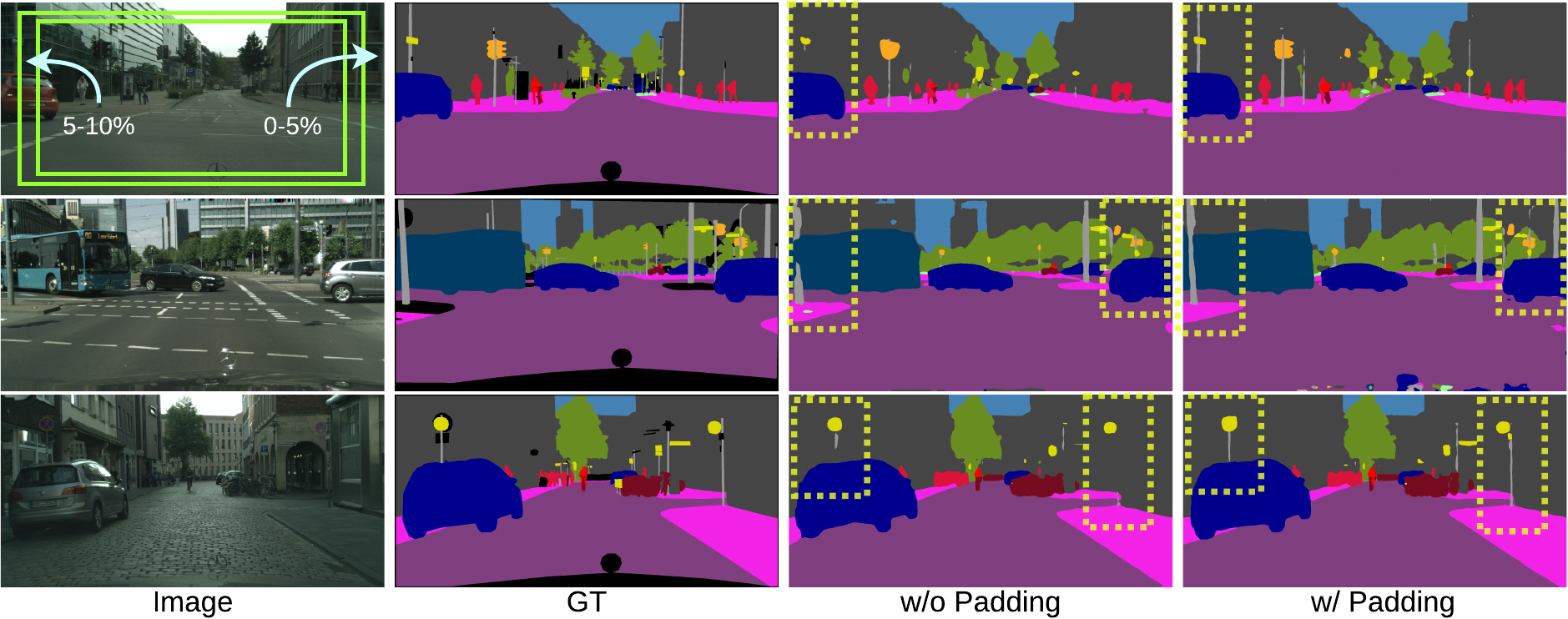}
          
			}
	\end{center}
	\vspace{-0.35cm}
	\caption{Example predictions on the Cityscapes validation set when training with and without \textit{padding}. Best viewed zoomed in.}
	
	\label{fig:city}
\vspace{-0.1cm}
\end{figure*}

\section{Applicability to Semantic Segmentation, Texture Recognition, Data Augmentation, and Adversarial Robustness}\label{sec:real}
Given the intriguing findings above, it is natural to ask how much the demonstrated phenomenon affects real world tasks with SOTA architectures. More specifically, does encoding position always improve performance or does it cause unwanted effects on certain tasks (\textbf{H-V})?  

\subsection{Semantic Segmentation} We now measure the impact of zero padding to segment objects near the image boundary with a strong semantic segmentation network on an automotive-centric dataset. We use the DeepLabv3~\cite{chen2017rethinking} network and the Cityscapes~\cite{cordts2016cityscapes} dataset, trained with different padding types. From Table~\ref{tab:seg_city}, it is clear that DeepLabv3 with zero padding achieves superior results compared to the model trained without padding or with reflect padding. Additionally, we perform an analysis by computing the mIoU for rectangular ring-like regions (see Fig.~\ref{fig:city} (top-left)), between $X\%$ and $Y\%$, where $X$ and $Y$ are relative distances from the border (e.g., $0\%-5\%$ is the outer most region of the image, while $5\%-10\%$ is the neighbouring inner $5\%$ region) to quantify the performance decrease from the boundary effect and lack of positional information. From Table~\ref{tab:seg_city}, the performance drop between the total mIoU (100\%) and the border region (0-5\%) is more significant for the no padding case and reflect padding case compared to the zero padding case, which agrees with the results found in Sec.~\ref{sec:analysis}. This further demonstrates that the absolute position information due to zero padding improves the performance at all image regions, while reflect padding is not as beneficial at the image boundaries. Figure~\ref{fig:city} shows examples of how DeepLabv3 trained with zero padding generates more accurate predictions, particularly near the border of the image. Note that thin or complex objects near the border regions are particularly affected (e.g., light posts). The reason that performance suffers even \textit{with} padding, is the lack of semantic and contextual information near the border, which is not the case for grid-based tasks (Sec.~\ref{sec:loc_class_seg}) since the image patch contains the entire CIFAR-10 image. Additional results can be found in Sec. A.5 in the Appendix.

				  
	

\subsection{Texture Recognition} We evaluate three models with three padding types on the task of texture recognition. We use a ResNet-34, ResNet-50, and VGG-5 trained with zero, reflect, and no padding settings, with the GTOS-Mobile dataset~\cite{xue2018deep} and Describable Textures Dataset (DTD)~\cite{cimpoi2014describing}. We hypothesize that, since there is little to no spatial bias (e.g., orientation) in most texture recognition datasets, position information may not benefit the performance of the CNN. As shown in Table~\ref{tab:texture}, models trained with reflect padding outperform the models trained with zero padding. This result implies that position information may not guide the network to learn robust representations for the task of texture recognition. Note that, although no padding has less position information than reflect padding, the CNN suffers from the border effects without padding (see Fig.~\ref{fig:border_activation}), which hurts performance significantly (i.e., since the kernel's support does not cover the entire image domain). 

\begin{table} [t]
\caption{\text{Texture recognition} results on two datasets with different padding types. Interestingly, reflect padding outperforms zero padding for the texture recognition task.}
     \centering
     
       \setlength\tabcolsep{4.1pt}
		\def\arraystretch{1.2}
    \resizebox{0.48\textwidth}{!}{
         \begin{tabular}{l|ccc|ccc}
				 \specialrule{1.2pt}{1pt}{1pt}
				 \multirow{2}{*}{Padding} & \multicolumn{3}{c|}{GTOS-M} & \multicolumn{3}{c}{DTD} \\
				 \cline{2-7}
				   & Res34 & Res50 & VGG5 & Res34 & Res50 & VGG5\\
				  \specialrule{1.2pt}{1pt}{1pt}
				  No Pad & 71.7 &  76.3& 33.6 & 57.5& 67.0& 27.3\\
				  Zero Pad & 78.7 & 81.7 & 39.7 &68.6 & 70.6& 32.8\\
				  \textbf{Reflect} & \textbf{80.6} & \textbf{85.0} & \textbf{43.1} & \textbf{70.4}& \textbf{71.7}& \textbf{34.0}\\
				  				\specialrule{1.2pt}{1pt}{1pt}

			\end{tabular}
	}
	
\label{tab:texture}
\end{table} 
\begin{table}
\caption{\text{Performance and robustness} of DeepLabv3 variants trained with Cutout~\cite{devries2017improved} using two canvas (Black and White) settings. Clearly, DeepLabv3 variants trained with white canvas based Cutout is more robust to the adversarial attacks than the black canvas based Cutout.}
     \centering
     
    \setlength\tabcolsep{4.1pt}
	\def\arraystretch{1.2}
    \resizebox{0.45\textwidth}{!}{
		\begin{tabular}{l|cc|cc}
				\specialrule{1.2pt}{1pt}{1pt}
				\multirow{2}{*}{\hspace{0.08cm} Method}& \multicolumn{2}{c|}{Segmentation} & 
				\multicolumn{2}{c}{Robustness}  \\
				\cline{2-5}

				& \textbf{B} & \textbf{W} & \textbf{B} & \textbf{W} \\
			    \specialrule{1.2pt}{1pt}{1pt}
			    
			    DeepLabv3-Res50 & 73.9&74.1& 53.7 & \textbf{55.8}\\
			    DeepLabv3-Res101 & 75.5 & 75.2 & 49.8 & \textbf{51.9}\\
			    \specialrule{1.2pt}{1pt}{1pt}
			    

							\end{tabular}
	}
	
\label{tab:canvas}
\end{table} 

\subsection{Canvas Analysis: Cutout \& Adversarial Robustness} We investigate the impact of different canvas colors in terms of performance and robustness using a data augmentation strategy, Cutout~\cite{devries2017improved}, which simply places a rectangular black mask over random image regions during training. We evaluate DeepLabv3 with two backbones using the Cutout strategy for semantic segmentation on the PASCAL VOC 2012~\cite{PASCALVOC} dataset with black \textit{and} white masks (see Fig. 20 in the appendix for example inputs). We also evaluate the robustness of each model to show which canvas is more resilient to the GD-UAP adversarial attack~\cite{mopuri2018generalizable}. Note that the GD-UAP attack is generated based on the image-agnostic DeepLab-ResNet101 backbone. As shown in Table~\ref{tab:canvas}, DeepLabv3 trained with white-mask Cutout is significantly more robust to adversarial examples than the black canvas, without sacrificing segmentation performance.



     
				

                
                
                
                
                
                
                
			

\section{Conclusion}\label{Conclusion}
In this paper, we first explored the hypothesis that absolute position information is implicitly encoded in convolutional neural networks. Experiments reveal that positional information is available to a strong degree.
Results point to zero padding and borders as an anchor from which spatial information is derived and eventually propagated over the whole image as spatial abstraction occurs.
Further, with the goal of answering whether boundary effects are a feature or a bug, we have presented evidence that the heuristics used at the image boundary play a much deeper role in a CNN's ability to perform different tasks than one might assume. By designing a series of location dependant experiments, we have performed a unique exploration into how this connection reveals itself. We showed that zero padding encodes more position information relative to common padding types (\textbf{H-I}) and that zero padding causes more dimensions to encode position information and that this correlates with the number of dimensions that encode semantics (\textbf{H-III}). We examined the ability of CNNs to perform semantic tasks as a function of the distance to a border. This revealed the capability of a \textit{black} canvas to provide rich position information compared to other colors (i.e., \textit{White} and \textit{Mean}) (\textbf{H-II}). We visualized a number of features in CNNs which showed that boundary effects have an impact on \textit{all} regions of the input (\textbf{H-IV}), and highlighted characteristics of border handling techniques which allow for absolute position information to be encoded. This position encoding enables CNNs to learn more separable semantic features which provide more accurate and confident predictions (\textbf{H-III}). We conducted these experiments with the following question in mind: Are boundary effects a feature or a bug (\textbf{H-V})? After teasing out the above underlying properties, we were able to validate the hypothesis that different types of padding, levels of position information, and canvas colors, could be beneficial \textit{depending on the task at hand}! To be more clear: the position information can be used to improve performance, but can also be detrimental to a CNNs performance if not taken into consideration. These results demonstrate a fundamental property of CNNs that was unknown to date, and for which much further exploration is warranted.

\ifCLASSOPTIONcaptionsoff
  \newpage
\fi



%
\bibliographystyle{IEEEtran}
\bibliography{./position}

\begin{thebibliography}{10}
\providecommand{\url}[1]{#1}
\csname url@samestyle\endcsname
\providecommand{\newblock}{\relax}
\providecommand{\bibinfo}[2]{#2}
\providecommand{\BIBentrySTDinterwordspacing}{\spaceskip=0pt\relax}
\providecommand{\BIBentryALTinterwordstretchfactor}{4}
\providecommand{\BIBentryALTinterwordspacing}{\spaceskip=\fontdimen2\font plus
\BIBentryALTinterwordstretchfactor\fontdimen3\font minus
  \fontdimen4\font\relax}
\providecommand{\BIBforeignlanguage}[2]{{%
\expandafter\ifx\csname l@#1\endcsname\relax
\typeout{** WARNING: IEEEtran.bst: No hyphenation pattern has been}%
\typeout{** loaded for the language `#1'. Using the pattern for}%
\typeout{** the default language instead.}%
\else
\language=\csname l@#1\endcsname
\fi
#2}}
\providecommand{\BIBdecl}{\relax}
\BIBdecl

\bibitem{krizhevsky2012imagenet}
A.~Krizhevsky, I.~Sutskever, and G.~E. Hinton, ``Image{N}et classification with
  deep convolutional neural networks,'' in \emph{NIPS}, 2012.

\bibitem{Simonyan14}
K.~Simonyan and A.~Zisserman, ``Very deep convolutional networks for
  large-scale image recognition,'' in \emph{ICLR}, 2015.

\bibitem{szegedy2015going}
C.~Szegedy, W.~Liu, Y.~Jia, P.~Sermanet, S.~Reed, D.~Anguelov, D.~Erhan,
  V.~Vanhoucke, and A.~Rabinovich, ``Going deeper with convolutions,'' in
  \emph{CVPR}, 2015.

\bibitem{huang2017densely}
G.~Huang, Z.~Liu, L.~Van Der~Maaten, and K.~Q. Weinberger, ``Densely connected
  convolutional networks,'' in \emph{CVPR}, 2017.

\bibitem{karpathy2014large}
A.~Karpathy, G.~Toderici, S.~Shetty, T.~Leung, R.~Sukthankar, and L.~Fei-Fei,
  ``Large-scale video classification with convolutional neural networks,'' in
  \emph{CVPR}, 2014.

\bibitem{yue2015beyond}
J.~Yue-Hei~Ng, M.~Hausknecht, S.~Vijayanarasimhan, O.~Vinyals, R.~Monga, and
  G.~Toderici, ``Beyond short snippets: Deep networks for video
  classification,'' in \emph{CVPR}, 2015.

\bibitem{carreira2017quo}
J.~Carreira and A.~Zisserman, ``Quo vadis, action recognition? {A} new model
  and the kinetics dataset,'' in \emph{CVPR}, 2017.

\bibitem{Ren15}
S.~Ren, K.~He, R.~Girshick, and J.~Sun, ``Faster {R-CNN}: Towards real-time
  object detection with region proposal networks,'' in \emph{NIPS}, 2015.

\bibitem{Redmon15}
J.~Redmon, S.~Divvala, R.~Girshick, and A.~Farhadi, ``You only look once:
  Unified, real-time object detection,'' in \emph{CVPR}, 2016.

\bibitem{he2017mask}
K.~He, G.~Gkioxari, P.~Doll{\'a}r, and R.~Girshick, ``Mask {R-CNN},'' in
  \emph{ICCV}, 2017.

\bibitem{brock2018large}
A.~Brock, J.~Donahue, and K.~Simonyan, ``Large {s}cale {GAN} training for high
  fidelity natural image synthesis,'' \emph{arXiv:1809.11096}, 2018.

\bibitem{long15_cvpr}
J.~Long, E.~Shelhamer, and T.~Darrell, ``Fully convolutional networks for
  semantic segmentation,'' in \emph{CVPR}, 2015.

\bibitem{noh15_iccv}
H.~Noh, S.~Hong, and B.~Han, ``Learning deconvolution network for semantic
  segmentation,'' in \emph{ICCV}, 2015.

\bibitem{islam2017label}
M.~A. Islam, S.~Naha, M.~Rochan, N.~Bruce, and Y.~Wang, ``Label refinement
  network for coarse-to-fine semantic segmentation,'' \emph{arXiv:1703.00551},
  2017.

\bibitem{islam2017gated}
M.~A. Islam, M.~Rochan, N.~D. Bruce, and Y.~Wang, ``Gated feedback refinement
  network for dense image labeling,'' in \emph{CVPR}, 2017.

\bibitem{chen2017rethinking}
L.-C. Chen, G.~Papandreou, F.~Schroff, and H.~Adam, ``Rethinking atrous
  convolution for semantic image segmentation,'' \emph{arXiv:1706.05587}, 2017.

\bibitem{islam2018gated}
M.~A. Islam, M.~Rochan, S.~Naha, N.~D. Bruce, and Y.~Wang, ``Gated feedback
  refinement network for coarse-to-fine dense semantic image labeling,''
  \emph{arXiv preprint arXiv:1806.11266}, 2018.

\bibitem{karim2019recurrent}
R.~Karim, M.~A. Islam, and N.~D. Bruce, ``Recurrent iterative gating networks
  for semantic segmentation,'' in \emph{WACV}, 2019.

\bibitem{chen2018deeplab}
L.-C. Chen, G.~Papandreou, I.~Kokkinos, K.~Murphy, and A.~L. Yuille, ``Deeplab:
  Semantic image segmentation with deep convolutional nets, atrous convolution,
  and fully connected {CRFs},'' \emph{TPAMI}, 2018.

\bibitem{karim2020distributed}
R.~Karim, M.~A. Islam, and N.~D. Bruce, ``Distributed iterative gating networks
  for semantic segmentation,'' in \emph{WACV}, 2020.

\bibitem{islam2020sbinding}
M.~A. Islam, M.~Kowal, K.~G. Derpanis, and N.~D. Bruce, ``Feature binding with
  category-dependant mixup for semantic segmentation and adversarial
  robustness,'' in \emph{BMVC}, 2020.

\bibitem{Liu_2016_CVPR}
N.~Liu and J.~Han, ``Dhsnet: Deep hierarchical saliency network for salient
  object detection,'' in \emph{CVPR}, 2016.

\bibitem{islam2017salient}
M.~A. Islam, M.~Kalash, M.~Rochan, N.~D. Bruce, and Y.~Wang, ``Salient object
  detection using a context-aware refinement network,'' in \emph{BMVC}, 2017.

\bibitem{cvpr18_rank}
M.~A. Islam, M.~Kalash, and N.~D. Bruce, ``Revisiting salient object detection:
  Simultaneous detection, ranking, and subitizing of multiple salient
  objects,'' in \emph{CVPR}, 2018.

\bibitem{PICA}
N.~Liu, J.~Han, and M.-H. Yang, ``Picanet: Learning pixel-wise contextual
  attention for saliency detection,'' in \emph{CVPR}, 2018.

\bibitem{islam2018semantics}
M.~A. Islam, M.~Kalash, and N.~D. Bruce, ``Semantics meet saliency: Exploring
  domain affinity and models for dual-task prediction,'' in \emph{BMVC}, 2018.

\bibitem{kalash2019relative}
M.~Kalash, M.~A. Islam, and N.~D. Bruce, ``Relative saliency and ranking:
  Models, metrics, data and benchmarks,'' \emph{TPAMI}, 2019.

\bibitem{Jia18}
S.~Jia and N.~D. Bruce, ``Eml-net: An expandable multi-layer network for
  saliency prediction,'' \emph{arXiv:1805.01047}, 2018.

\bibitem{wohlberg2017convolutional}
B.~Wohlberg and P.~Rodriguez, ``Convolutional sparse coding: Boundary handling
  revisited,'' \emph{arXiv:1707.06718}, 2017.

\bibitem{tang2018high}
M.~Tang, L.~Zheng, B.~Yu, and J.~Wang, ``High speed kernelized correlation
  filters without boundary effect,'' \emph{arXiv:1806.06406}, 2018.

\bibitem{liu2018partial}
G.~Liu, K.~J. Shih, T.-C. Wang, F.~A. Reda, K.~Sapra, Z.~Yu, A.~Tao, and
  B.~Catanzaro, ``Partial convolution based padding,'' \emph{arXiv:1811.11718},
  2018.

\bibitem{innamorati2019learning}
C.~Innamorati, T.~Ritschel, T.~Weyrich, and N.~J. Mitra, ``Learning on the
  edge: Investigating boundary filters in {CNN}s,'' \emph{IJCV}, 2019.

\bibitem{liu2018intriguing}
R.~Liu, J.~Lehman, P.~Molino, F.~P. Such, E.~Frank, A.~Sergeev, and
  J.~Yosinski, ``An intriguing failing of convolutional neural networks and the
  coordconv solution,'' in \emph{NeurIPS}, 2018.

\bibitem{perez2019turing}
J.~P{\'e}rez, J.~Marinkovi{\'c}, and P.~Barcel{\'o}, ``On the {t}uring
  completeness of modern neural network architectures,'' in \emph{ICLR}, 2019.

\bibitem{kayhan2020translation}
O.~S. Kayhan and J.~C.~v. Gemert, ``On translation invariance in cnns:
  Convolutional layers can exploit absolute spatial location,'' in \emph{CVPR},
  2020.

\bibitem{gregor2015draw}
K.~Gregor, I.~Danihelka, A.~Graves, D.~J. Rezende, and D.~Wierstra, ``{DRAW}: A
  recurrent neural network for image generation,'' in \emph{ICML}, 2015.

\bibitem{huang2019learning}
Z.~Huang, W.~Heng, and S.~Zhou, ``Learning to paint with model-based deep
  reinforcement learning,'' in \emph{ICCV}, 2019.

\bibitem{devries2017improved}
T.~DeVries and G.~W. Taylor, ``Improved regularization of convolutional neural
  networks with cutout,'' \emph{arXiv preprint arXiv:1708.04552}, 2017.

\bibitem{demir2018patch}
U.~Demir and G.~Unal, ``Patch-based image inpainting with generative
  adversarial networks,'' \emph{arXiv preprint arXiv:1803.07422}, 2018.

\bibitem{yu2018generative}
J.~Yu, Z.~Lin, J.~Yang, X.~Shen, X.~Lu, and T.~S. Huang, ``Generative image
  inpainting with contextual attention,'' in \emph{CVPR}, 2018.

\bibitem{geirhos2018imagenet}
R.~Geirhos, P.~Rubisch, C.~Michaelis, M.~Bethge, F.~A. Wichmann, and
  W.~Brendel, ``Imagenet-trained {CNNs} are biased towards texture; increasing
  shape bias improves accuracy and robustness,'' in \emph{ICLR}, 2018.

\bibitem{esser2020disentangling}
P.~Esser, R.~Rombach, and B.~Ommer, ``A disentangling invertible interpretation
  network for explaining latent representations,'' in \emph{CVPR}, 2020.

\bibitem{he2016deep}
K.~He, X.~Zhang, S.~Ren, and J.~Sun, ``Deep residual learning for image
  recognition,'' in \emph{CVPR}, 2016.

\bibitem{tsotsos1995modeling}
J.~K. Tsotsos, S.~M. Culhane, W.~Y.~K. Wai, Y.~Lai, N.~Davis, and F.~Nuflo,
  ``Modeling visual attention via selective tuning,'' \emph{Artificial
  intelligence}, 1995.

\bibitem{islam2020much}
M.~A. Islam, S.~Jia, and N.~D. Bruce, ``How much position information do
  convolutional neural networks encode?'' in \emph{ICLR}, 2020.

\bibitem{Zeiler11}
M.~D. {Zeiler}, G.~W. {Taylor}, and R.~{Fergus}, ``Adaptive deconvolutional
  networks for mid and high level feature learning,'' in \emph{ICCV}, 2011.

\bibitem{ZeilerF13}
M.~D. Zeiler and R.~Fergus, ``Visualizing and understanding convolutional
  networks,'' in \emph{ECCV}, 2014.

\bibitem{Zhang16}
C.~Zhang, S.~Bengio, M.~Hardt, B.~Recht, and O.~Vinyals, ``Understanding deep
  learning requires rethinking generalization,'' \emph{arXiv:1611.03530}, 2016.

\bibitem{zhou2016learning}
B.~Zhou, A.~Khosla, A.~Lapedriza, A.~Oliva, and A.~Torralba, ``Learning deep
  features for discriminative localization,'' in \emph{CVPR}, 2016.

\bibitem{Selvaraju16}
R.~R. Selvaraju, M.~Cogswell, A.~Das, R.~Vedantam, D.~Parikh, and D.~Batra,
  ``Grad-{CAM}: Visual explanations from deep networks via gradient-based
  localization,'' in \emph{ICCV}, 2017.

\bibitem{denton2017unsupervised}
E.~L. Denton \emph{et~al.}, ``Unsupervised learning of disentangled
  representations from video,'' in \emph{NIPS}, 2017.

\bibitem{lorenz2019unsupervised}
D.~Lorenz, L.~Bereska, T.~Milbich, and B.~Ommer, ``Unsupervised {P}art-based
  disentangling of object shape and appearance,'' in \emph{CVPR}, 2019.

\bibitem{alsallakh2020mind}
B.~Alsallakh, N.~Kokhlikyan, V.~Miglani, J.~Yuan, and O.~Reblitz-Richardson,
  ``Mind the pad--{CNNs} can develop blind spots,'' \emph{arXiv preprint
  arXiv:2010.02178}, 2020.

\bibitem{xu2020positional}
R.~Xu, X.~Wang, K.~Chen, B.~Zhou, and C.~C. Loy, ``Positional encoding as
  spatial inductive bias in {GANs},'' \emph{arXiv preprint arXiv:2012.05217},
  2020.

\bibitem{wang2018location}
Z.~Wang and O.~Veksler, ``Location augmentation for {CNN},''
  \emph{arXiv:1807.07044}, 2018.

\bibitem{murase2020can}
R.~Murase, M.~Suganuma, and T.~Okatani, ``How can {CNNs} use image position for
  segmentation?'' \emph{arXiv:2005.03463}, 2020.

\bibitem{Sabour17}
S.~Sabour, N.~Frosst, and G.~E. Hinton, ``Dynamic routing between capsules,''
  in \emph{NIPS}, 2017.

\bibitem{Visin15}
F.~Visin, K.~Kastner, K.~Cho, M.~Matteucci, A.~Courville, and Y.~Bengio, ``A
  recurrent neural network based alternative to convolutional networks,''
  \emph{arXiv:1505.00393}, 2015.

\bibitem{sirovich1979effect}
L.~Sirovich, S.~E. Brodie, and B.~Knight, ``Effect of boundaries on the
  response of a neural network,'' \emph{Biophysical}, 1979.

\bibitem{le2015tiny}
Y.~Le and X.~Yang, ``Tiny {ImageNet} visual recognition challenge,'' \emph{CS
  231N}, 2015.

\bibitem{DUTS}
L.~Wang, H.~Lu, Y.~Wang, M.~Feng, D.~Wang, B.~Yin, and X.~Ruan, ``Learning to
  detect salient objects with image-level supervision,'' in \emph{CVPR}, 2017.

\bibitem{PASCALS}
Y.~Li, X.~Hou, C.~Koch, J.~M. Rehg, and A.~L. Yuille, ``The secrets of salient
  object segmentation,'' in \emph{CVPR}, 2014.

\bibitem{krizhevsky2014cifar}
A.~Krizhevsky, V.~Nair, and G.~Hinton, ``The {CIFAR}-10 dataset,''
  \emph{online: http://www. cs. toronto. edu/kriz/cifar. html}, 2014.

\bibitem{brendel2019approximating}
W.~Brendel and M.~Bethge, ``Approximating {CNNs} with bag-of-local-features
  models works surprisingly well on {ImageNet},'' in \emph{ICLR}, 2019.

\bibitem{bau2017network}
D.~Bau, B.~Zhou, A.~Khosla, A.~Oliva, and A.~Torralba, ``Network dissection:
  Quantifying interpretability of deep visual representations,'' in
  \emph{CVPR}, 2017.

\bibitem{islam2020shape}
M.~A. Islam, M.~Kowal, P.~Esser, S.~Jia, B.~Ommer, K.~G. Derpanis, and
  N.~Bruce, ``Shape or texture: Understanding discriminative features in
  {CNNs},'' in \emph{ICLR}, 2021.

\bibitem{maaten2008visualizing}
L.~v.~d. Maaten and G.~Hinton, ``Visualizing data using t-{SNE},'' \emph{JMLR},
  2008.

\bibitem{cordts2016cityscapes}
M.~Cordts, M.~Omran, S.~Ramos, T.~Rehfeld, M.~Enzweiler, R.~Benenson,
  U.~Franke, S.~Roth, and B.~Schiele, ``The {C}ityscapes dataset for semantic
  urban scene understanding,'' in \emph{CVPR}, 2016.

\bibitem{xue2018deep}
J.~Xue, H.~Zhang, and K.~Dana, ``Deep texture manifold for ground terrain
  recognition,'' in \emph{CVPR}, 2018.

\bibitem{cimpoi2014describing}
M.~Cimpoi, S.~Maji, I.~Kokkinos, S.~Mohamed, and A.~Vedaldi, ``Describing
  textures in the wild,'' in \emph{CVPR}, 2014.

\bibitem{PASCALVOC}
M.~Everingham, L.~Van~Gool, C.~K.~I. Williams, J.~Winn, and A.~Zisserman, ``The
  {PASCAL} {V}isual {O}bject {C}lasses {C}hallenge 2010 {(VOC2010)}
  {R}esults,''
  http://www.pascal-network.org/challenges/VOC/voc2010/workshop/index.html,
  2010.

\bibitem{mopuri2018generalizable}
K.~R. Mopuri, A.~Ganeshan, and R.~V. Babu, ``Generalizable data-free objective
  for crafting universal adversarial perturbations,'' \emph{TPAMI}, 2018.

\bibitem{AMU}
P.~Zhang, D.~Wang, H.~Lu, H.~Wang, and X.~Ruan, ``Amulet: Aggregating
  multi-level convolutional features for salient object detection,'' in
  \emph{ICCV}, 2017.

\bibitem{Xavier}
X.~Glorot and Y.~Bengio, ``Understanding the difficulty of training deep
  feedforward neural networks,'' in \emph{AISTATS}, 2010.

\end{thebibliography}

%

\begin{IEEEbiography}[{\includegraphics[width=1in,height=1.25in,clip,keepaspectratio]{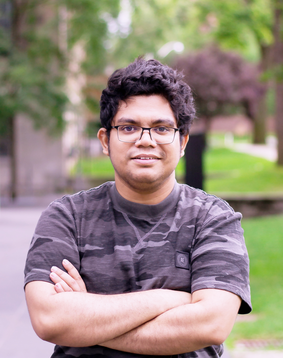}}]{Md Amirul Islam}
is currently a Ph.D. student at the Department of Computer Science at Ryerson University, Canada. He is also a Postgraduate Affiliate at the Vector Institute for AI, Toronto. He received his M.Sc. in Computer Science from University of Manitoba, Canada in 2017 and his B.Sc. in Computer Science and Engineering from North South University, Bangladesh in 2014. He has worked as a research intern at Noah’s Ark Labs, Huawei Canada, Toronto in summer 2019 and 2020. His research interests are in the area of computer vision, with a focus on exploring various mechanisms which allow for humans to understand the different properties of CNNs and semantic understanding of a scene.
\end{IEEEbiography}

\begin{IEEEbiography}[{\includegraphics[width=1in,height=1.25in,clip,keepaspectratio]{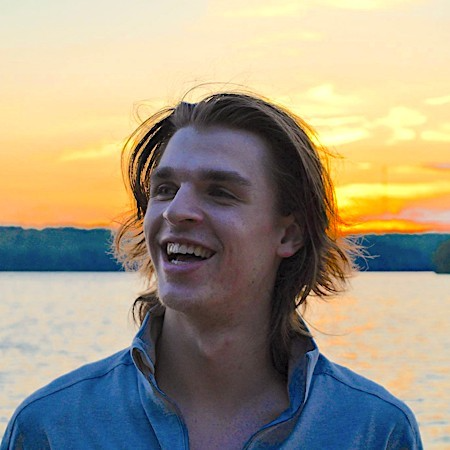}}]{Matthew Kowal}
received a B.A.Sc. in Applied Mathematics and Engineering from Queen's University, Canada in 2017, and a M.Sc. in Computer Science from Ryerson University, Canada in 2020. He is currently pursuing his Ph.D. in Computer Science from Ryerson University, Canada. In 2020, he joined NextAI as a Scientist in Residence. His research interests include computer vision and more specifically designing interpretable deep learning algorithms for various visual tasks.
\end{IEEEbiography}

\begin{IEEEbiography}[{\includegraphics[width=1in,height=1.25in,clip,keepaspectratio]{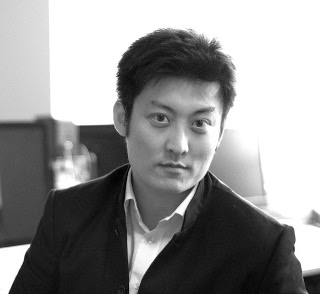}}]{Sen Jia}
is a postdoctoral researcher in the Vision and Image Processing lab of University of Waterloo. He has a wide range of research areas, including saliency detection in computer vision and uncertainty in neural networks. Prior to this, he worked as a postdoctoral researcher at Ryerson University(Canada) and Bristol University (UK) in 2018 and 2016 respectively. He received his PhD degree from the University of Bristol in 2017 under the supervision of Prof. Nello Cristianini. He received his Master of Science (Msc) degree with distinction from the University of Newcastle in 2010 and Bachelor of Engineering (BE) from Beijing University of Technology in 2008.
\end{IEEEbiography}

\begin{IEEEbiography}[{\includegraphics[width=1in,height=1.25in,clip,keepaspectratio]{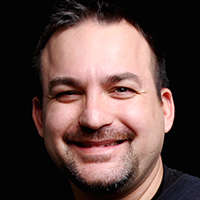}}]{Konstantinos G. Derpanis}
received the Honours Bachelor of Science (BSc) degree in computer science from the University of Toronto, Canada, in 2000, and the MSc (supervisors John Tsotsos and Richard Wildes) and PhD (supervisor Richard Wildes) degrees in computer science from York University, Canada, in 2003 and 2010, respectively. For his dissertation work, he received the Canadian Image Processing and Pattern Recognition Society (CIPPRS) Doctoral Dissertation Award 2010 Honourable Mention. Subsequently, he was a postdoctoral researcher in the GRASP Laboratory at the University of Pennsylvania under the supervision of Kostas Daniilidis. In 2012, he joined the Department of Computer Science at Ryerson University, Toronto, where he is an associate professor. He is a Faculty Affiliate at the Vector Institute for AI, Toronto. In 2019, Kosta joined the Samsung AI Centre in Toronto as a Research Scientist.
He currently serves as an AE for TPAMI and is an AC for CVPR 2021 and ICCV 2021.  His main research field of interest is computer vision with emphasis on motion analysis and human motion understanding, and related aspects in image processing and machine learning.
\end{IEEEbiography}

\begin{IEEEbiography}[{\includegraphics[width=1in,height=1.25in,clip,keepaspectratio]{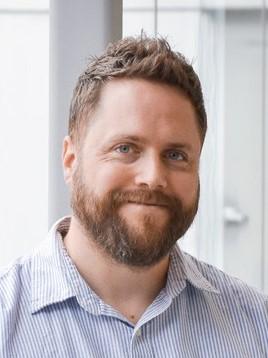}}]{Neil D. B. Bruce}
Dr. Neil Bruce graduated from the University of Guelph with a B.Sc. Double major in CS and Pure Mathematics. Dr. Bruce then attended the University of Waterloo for an M.A.Sc. in System Design Engineering and York University for a Ph.D. in Computer Science. Prior to joining Guelph he worked in the Department of Computer Science at Ryerson University. Prior to this Dr. Bruce worked at the University of Manitoba as Assistant then Associate Professor. Dr. Bruce has postdoctoral experience working at INRIA (France) and Epson Canada. He is the recipient of the Falconer Rh Young Researcher Award and is a Faculty Affiliate at the Vector Institute for AI, Toronto. His research has explored solutions to issues in computer vision, deep-learning, human perception, neuroscience and visual computing.
\end{IEEEbiography}






\clearpage
 \newpage

\appendices
\section{}
\subsection{Experimental Details of Absolute Position Encoding Experiments} \label{sec:exp_details}
\textbf{Datasets:} 
We use the DUT-S dataset \cite{DUTS} as our training set, which contains $10,533$ images for training. Following the common training protocol used in \cite{AMU,PICA}, we train the model on the training set of DUT-S and evaluate the existence of position information on the natural images of the PASCAL-S \cite{PASCALS} dataset. The synthetic images (white, black and Gaussian noise) are also used as described in Sec. 3.4 of the main manuscript. Note that we follow the common setting used in saliency detection just to make sure that there is no overlap between the training and test sets. However, any images can be used in our experiments given that the position information is relatively content independent.

\noindent \textbf{Evaluation Metrics:} As position encoding measurement is a new direction, there is no universal metric.  We use two different natural choices for metrics (Spearmen Correlation (SPC) and Mean Absoute Error (MAE)) to measure the position encoding performance. The SPC is defined as the Spearman's correlation between the ground-truth and the predicted position map. For ease of interpretation, we keep the SPC score within range [-1 1].
MAE is the average pixel-wise difference between the predicted position map and the ground-truth gradient position map.

\noindent \textbf{Implementation Details}\label{imp}
We initialize the architecture with a network pretrained for the ImageNet classification task. The new layers in the position encoding branch are initialized with \textit{xavier initialization} \cite{Xavier}. We train the networks using stochastic gradient descent for 15 epochs with momentum of $0.9$, and weight decay of $1e-4$. We resize each image to a fixed size of $224 \times 224$ during training and inference. Since the spatial extent of multi-level features are different, we align all the feature maps to a size of $28 \times 28$. 

\subsection{Implementation Deatils of VGG-5 Network for Position Information} \label{sec:vgg}
We use a simplified VGG network (VGG-5) for the position encoding experiments in Sec. 3.4 of the main manuscript and texture recognition experiments in Sec. 7 of the main manuscript. The details of the VGG-5 architecture are shown in Table~\ref{tab:vgg5} (in this table we show the VGG-5 network trained on the tiny ImageNet dataset, the VGG-5 network trained on texture recognition has a different input size: $224 \times 224$). Note that the network is trained from scratch. The tiny ImageNet dataset contains 200 classes and each class has 500 images for training and 50 for validation. The size of the input image is $64 \times 64$, a random crop of $56 \times 56$ is used for training and a center crop is applied for validation. The total training epochs is set to 100 with an initial learning rate of $0.01$. The learning rate was decayed at the $60th$ and $80th$ epochs by multiplying the learning rate by a factor of $0.1$. A momentum of $0.9$ and a weight decay of $1e-4$ are applied with the the stochastic gradient descent optimizer. After the pre-training process, a simple read-out module is applied on the pre-trained frozen backbone for position evaluation, following the training protocol as used in~\cite{islam2020much}. Note that the type of padding strategy is consistent between the pre-training and position evaluation procedures.
\begin{table} [t]
\caption{Configuration of VGG-5 architecture trained on tiny ImageNet.}
\def\arraystretch{1.25}
\centering

\resizebox{0.49\textwidth}{!}{
	\begin{tabular}{c }
	\specialrule{1.2pt}{1pt}{1pt}
	 
	 RGB image $x\in \mathbb{R}^{56\times56\times3}$ \\
	 \midrule
	 Conv2d ($3\times3$), Batch Norm, ReLU, MaxPool2d $\rightarrow$ $\mathbb{R}^{28\times28\times32}$ \\
	  \midrule
	  Conv2d ($3\times3$), Batch Norm, ReLU, MaxPool2d $\rightarrow$ $\mathbb{R}^{14\times14\times64}$ \\
	   \midrule
	  Conv2d ($3\times3$), Batch Norm, ReLU, MaxPool2d $\rightarrow$ $\mathbb{R}^{7\times7\times128}$ \\
	   \midrule
	  Conv2d ($3\times3$), Batch Norm, ReLU $\rightarrow$ $\mathbb{R}^{7\times7\times256}$ \\
	   \midrule
	  Global Average Pooling (GAP) $\rightarrow$ $\mathbb{R}^{1\times1\times256}$ \\
	   \midrule
	  FC $\rightarrow$ (256, classes) \\

	\specialrule{1.2pt}{1pt}{1pt}
	
	

		\end{tabular}
	
	}
	
    \label{tab:vgg5}
\end{table}
\begin{table*}[t]
	\caption{\textbf{Location dependant image segmentation:} Category-wise mIoU on CIFAR-10~\cite{krizhevsky2014cifar} for two different locations under w/ and w/o padding settings and Black and Mean canvas color. The grid size for both canvases is $7\times7$. Clearly, the encoding of absolute position information, by means of zero padding or a black canvas, has a significant effect on a CNN’s ability to segment object by learning distinctive semantic features.}
\centering
\def\arraystretch{1.1}
     
       \setlength\tabcolsep{4.1pt}
		\def\arraystretch{1.25}
  \resizebox{0.85\textwidth}{!}{
	\begin{tabular}{l||cc|cc | cc|cc}
		\specialrule{1.2pt}{1pt}{1pt}\
		
		
		\multirow{3}{*}{Categories}&  
		\multicolumn{4}{c|}{\textbf{Black}} &  
		\multicolumn{4}{c}{\textbf{Mean}} \\
		\cline{2-9}
		&  \multicolumn{2}{c|}{\textbf{$L=7$}} &  
		\multicolumn{2}{c|}{\textbf{$L=25$}} &  \multicolumn{2}{c|}{\textbf{$L=7$}} &  
		\multicolumn{2}{c}{\textbf{$L=25$}}\\
		\cline{2-9}
		
		& w/ Pad & w/o Pad & w/ Pad & w/o Pad & w/ Pad & w/o Pad & w/ Pad & w/o Pad \\
		\specialrule{1.2pt}{1pt}{1pt}\
        Background & 0.99& 0.99& 0.99& 0.99& 0.99 &0.99 &  0.99  & 0.99\\
	    Plane  &0.67& 0.65&0.65&0.70& 0.60 & 0.61 &  0.64 & 0.65\\
	    Car &0.80& 0.72&0.76& 0.79& 0.68 & 0.69&  0.74 & 0.76\\
	    Bird &0.57& 0.57&0.57& 0.61& 0.52 & 0.44 &  0.57 & 0.56\\
	    Cat  &0.46& 0.41&0.43&0.46&  0.42&  0.31 &  0.43 & 0.42\\
	    Deer &0.63&0.63&0.62& 0.65& 0.58&  0.53 &  0.60 & 0.61\\
	    Dog  &0.53&0.51&0.53&0.54& 0.49&  0.46 &  0.51 & 0.51\\
	    Frog &0.67&0.67&0.64& 0.72& 0.62& 0.64&  0.65 & 0.71\\
	    Horse &0.70&0.68&0.70& 0.71& 0.66 & 0.64&  0.68 & 0.70\\
	    Ship &0.78&0.67&0.74&0.74 & 0.68 & 0.73&  0.72 & 0.75\\
	    Truck  &0.74&0.66&0.71&0.73& 0.68 &  0.65&   0.72 & 0.72\\
	    \hline 
	    
	   \textbf{Overall} &\textbf{0.66}&\textbf{0.65}&\textbf{0.66}&\textbf{0.70}& \textbf{0.63} & \textbf{0.61}&  \textbf{0.66} & \textbf{0.67} \\

		\specialrule{1.2pt}{1pt}{1pt} 
	\end{tabular}
	
			}
		
		\label{tab:class_semantic}
\end{table*}
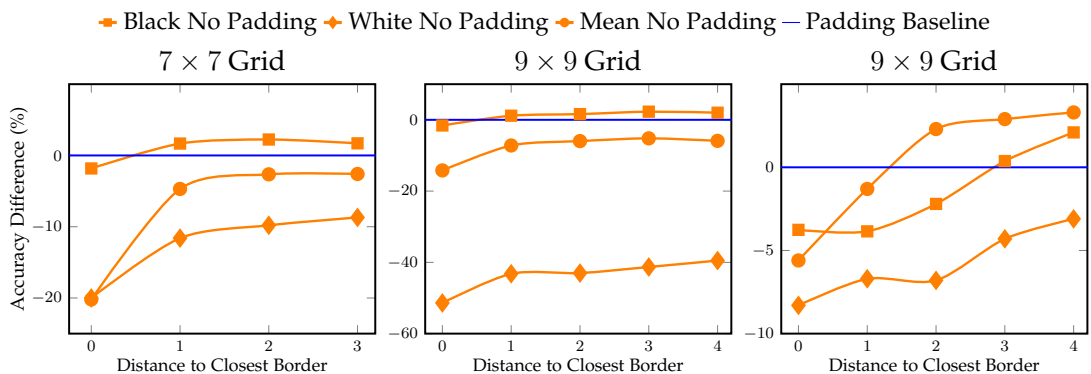
\begin{figure*}
	\begin{center}\ref{named_dist_diff_supp}
     \centering 
		\resizebox{0.8\textwidth}{!}{ \vspace{-0.7cm}
     \begin{tikzpicture}
    \begin{axis}[
       line width=1.0,
        title={$7\times 7$ Grid},
        title style={at={(axis description cs:0.5,1.12)},anchor=north,font=\Large},
        xlabel={Distance to Closest Border},
        ylabel={Accuracy Difference (\%)},
        xmin=-0.25, xmax=3.2,
        ymin=-25, ymax=10,
        xtick={0,1,2,3,4,5,6},
        ytick={0,-10,-20},
        x tick label style={font=\footnotesize},
        y tick label style={font=\footnotesize},
        x label style={at={(axis description cs:0.5,0.05)},anchor=north,font=\small},
        y label style={at={(axis description cs:0.12,.5)},anchor=south,font=\normalsize},
        width=7cm,
        height=6cm,        
        ymajorgrids=false,
        xmajorgrids=false,
        legend style={
         draw=none,
         nodes={scale=0.95, transform shape},
         cells={anchor=west},
         legend style={at={(-14.3,8.9)},anchor=north}},
         legend image post style={scale=0.4},
         legend columns=-1,
         legend entries={[black]Black No Padding,[black]White No Padding,[black]Mean No Padding, [black]Padding Baseline},
        legend to name=named_dist_diff_supp,
    ]
    \addplot[line width=1.2pt,mark size=2.5pt,smooth,color=orange,mark=square*,]
        coordinates {(0,-1.826)(1,1.66875)(2,2.2375)(3,1.7)};
    \addplot[line width=1.2pt,mark size=4pt,smooth,color=orange,mark=diamond*,]
        coordinates {(0,-20)(1,-11.6)(2,-9.8)(3,-8.7)};
    \addplot[line width=1.2pt,mark size=3pt,smooth,color=orange,mark=*,]
        coordinates {(0,-20.2)(1,-4.7)(2,-2.65)(3,-2.6)};
    \addplot[blue,sharp plot,update limits=false] 
	    coordinates {(-5,0)(10,0)};
    \end{axis}
\end{tikzpicture}
\begin{tikzpicture}
    \begin{axis}[
       line width=1.0,
        title={$9\times 9$ Grid},
        title style={at={(axis description cs:0.5,1.12)},anchor=north,font=\Large},
        xlabel={Distance to Closest Border},
        xmin=-0.25, xmax=4.2,
        ymin=-60, ymax=10,
        xtick={0,1,2,3,4,5,6},
        ytick={0,-20,-40,-60},
        x tick label style={font=\footnotesize},
        y tick label style={font=\footnotesize},
        x label style={at={(axis description cs:0.5,0.05)},anchor=north,font=\small},
        y label style={at={(axis description cs:0.12,.5)},anchor=south,font=\normalsize},
        width=7cm,
        height=6cm,        
        ymajorgrids=false,
        xmajorgrids=false,
    ]
    \addplot[line width=1.2pt,mark size=2.5pt,smooth,color=orange,mark=square*,]
        coordinates {(0,-1.5835)(1,1.15)(2,1.6)(3,2.2875)(4,2.0)};
    \addplot[line width=1.2pt,mark size=4pt,smooth,color=orange,mark=diamond*,]
        coordinates {(0,-51.4)(1,-43.2)(2,-43.0)(3,-41.3)(4,-39.5)};
    \addplot[line width=1.2pt,mark size=3pt,smooth,color=orange,mark=*,]
        coordinates {(0,-14.2)(1,-7.2)(2,-5.97)(3,-5.2)(4,-5.9)};
    \addplot[blue,sharp plot,update limits=false] 
	    coordinates {(-5,0)(10,0)};
    \end{axis}
\end{tikzpicture}
\begin{tikzpicture}
    \begin{axis}[
       line width=1.0,
        title={$9\times 9$ Grid},
        title style={at={(axis description cs:0.5,1.12)},anchor=north,font=\Large},
        xlabel={Distance to Closest Border},
        xmin=-0.25, xmax=4.2,
        ymin=-10, ymax=5,
        xtick={0,1,2,3,4,5,6},
        ytick={0,-5,-10},
        x tick label style={font=\footnotesize},
        y tick label style={font=\footnotesize},
        x label style={at={(axis description cs:0.5,0.05)},anchor=north,font=\small},
        y label style={at={(axis description cs:0.12,.5)},anchor=south,font=\normalsize},
        width=7cm,
        height=6cm,        
        ymajorgrids=false,
        xmajorgrids=false,
    ]
    \addplot[line width=1.2pt,mark size=2.5pt,smooth,color=orange,mark=square*,]
        coordinates {(0,-3.76)(1,-3.85)(2,-2.2)(3,0.375)(4,2.1)};
    \addplot[line width=1.2pt,mark size=4pt,smooth,color=orange,mark=diamond*,]
        coordinates {(0,-8.3)(1,-6.7)(2,-6.8)(3,-4.3)(4,-3.1)};
    \addplot[line width=1.2pt,mark size=3pt,smooth,color=orange,mark=*,]
        coordinates {(0,-5.6)(1,-1.3)(2,2.3)(3,2.9)(4,3.3)};
    \addplot[blue,sharp plot,update limits=false] 
	    coordinates {(-5,0)(10,0)};
    \end{axis}
\end{tikzpicture}
}
	\end{center}
	\vspace{-0.3cm}
	\caption{\textbf{Location dependant image classification (left two) and segmentation (right)}. Results show the accuracy difference between padding (blue horizontal line) and no padding (orange markers), at various distances to the border and canvas colors.}\label{fig:border_supp}
\end{figure*}

\subsection{Extended Per-location Analysis}
We now present additional `per-location' results. That is, we take advantage of the location dependant grid-based input and analyze the performance of CNNs at each location on the grid. This is done to reveal the impact of border effects with respect to the absolute location of the object of interest. We first show class-wise performance for the location dependant semantic segmentation task (Sec.~\ref{sec:class_seg}). Next, we show the performance as a function of the distance to the nearest border by averaging the accuracy over all locations which are a specified number of grid locations away from the nearest border (Sec.~\ref{sec:dist_to_border}). 
Note that all experiments are done with the same settings as Sec. 4 in the main paper, on the CIFAR-10~\cite{krizhevsky2014cifar} dataset. 
\begin{figure*}
	\begin{center}
		\includegraphics[width=0.99\textwidth]{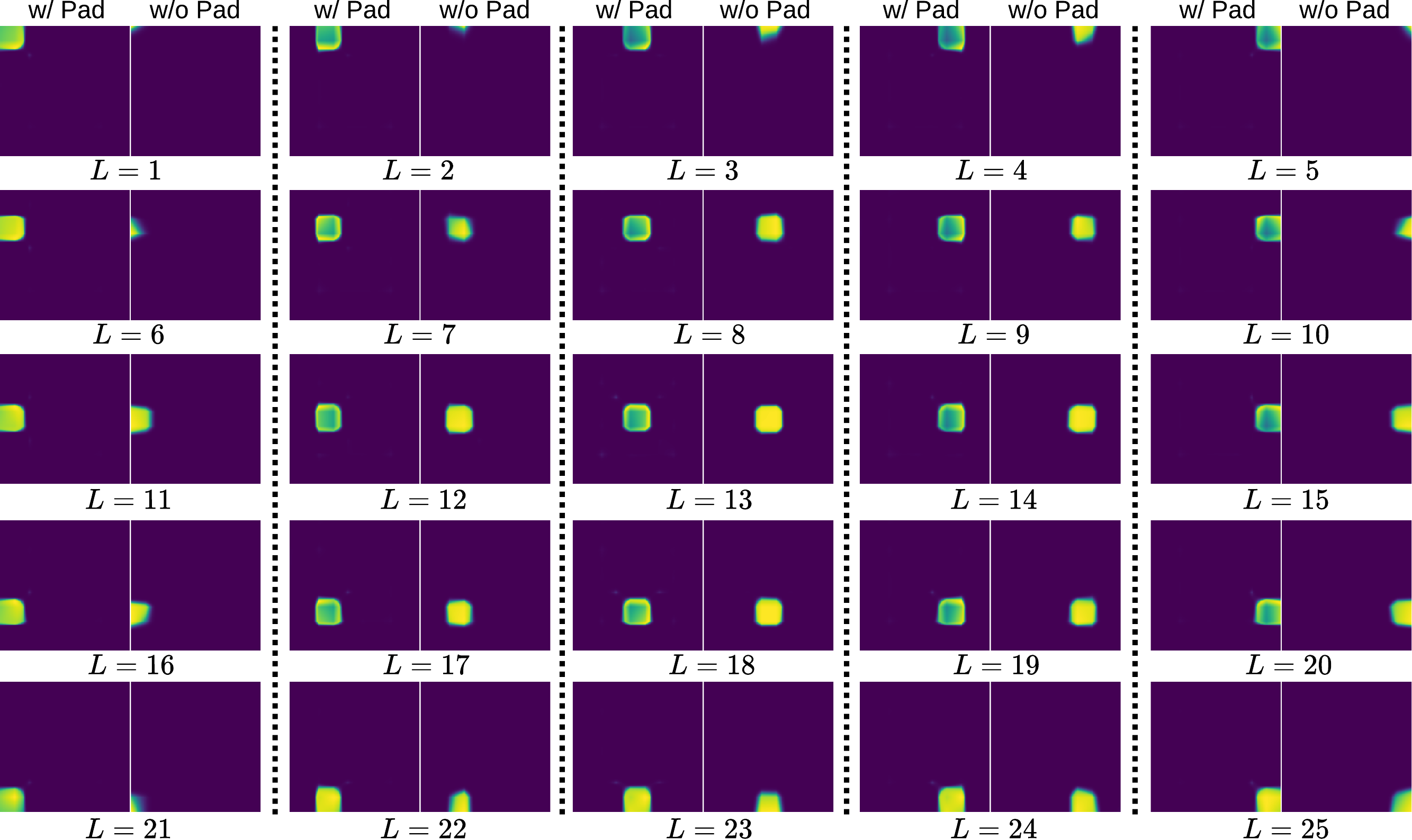}
		\caption{\textbf{Sample predictions of image segmentation} on all the locations of a $5\times5$ grid under the mean canvas setting. Confidence maps are plotted with the `viridis’ colormap,  where yellow and dark blue indicates higher and lower confidence, respectively.}
		\label{fig:semantic_preds}
	\end{center}
\end{figure*}
\subsubsection{Per-Location Category-wise {mIoU} Analysis}\label{sec:class_seg}
Table~\ref{tab:class_semantic} shows the category-wise mIoU for the location dependant image segmentation task for a $7\times7$ grid with black and mean canvas settings. We show the category-wise performance for a location at the very top right corner ($L=7$) and at the center of the grid ($L=25$), which highlights how the encoding of absolute position information affects the learning of semantic representations. 
For both locations, the border and the center, zero padding gives a large increase in performance for all classes compared to lack of padding. This is particularly pronounced with a mean canvas, demonstrating how the black canvas explicitly injects position information, even without the use of zero padding. For example, comparing 
the black and mean canvas at $L=7$ shows how important absolute position information can be in learning distinct semantic representations. The network trained with a mean canvas has a difficult time learning to segment images at this location when no padding is used and suffers a large drop in performance compared to the black canvas. Some classes even score around $1\%$ mIoU, which implies that the network fails to learn to segment certain classes (i.e., Bird, Cat, Deer, and Dog) with these settings. When zero padding is added (i.e., Mean, w/ padding, $L=7$), the network achieves a performance boost of between $35\%-60\%$. When a black canvas is used to inject position information instead (i.e., Black, w/o padding, $L=7$), the performance gains range from $15\%-40\%$.

Clearly, the encoding of position information, by means of zero padding or a black canvas, has a stark effect on a CNN's ability to learn distinctive semantic features. We see a similar, but not quite as drastic, pattern at the center of the image, further showing how the boundary effects impact all locations in an image, and not just at the image border.
\subsubsection{Distance to Border Performance}\label{sec:dist_to_border}
 Figure~\ref{fig:border_supp} shows the performance as a function of the distance to the closest border for all three canvas colors. The networks with zero padding are represented as a blue horizontal line, where the plotted markers show the difference in performance when no padding is used. Consistent with the results in the main paper, locations near the border are on average, much more difficult for networks to classify and segment, particularly as the grid size increases. 

\begin{figure*}
\centering
	\resizebox{0.85\textwidth}{!}{
          \includegraphics[width=0.8\textwidth]{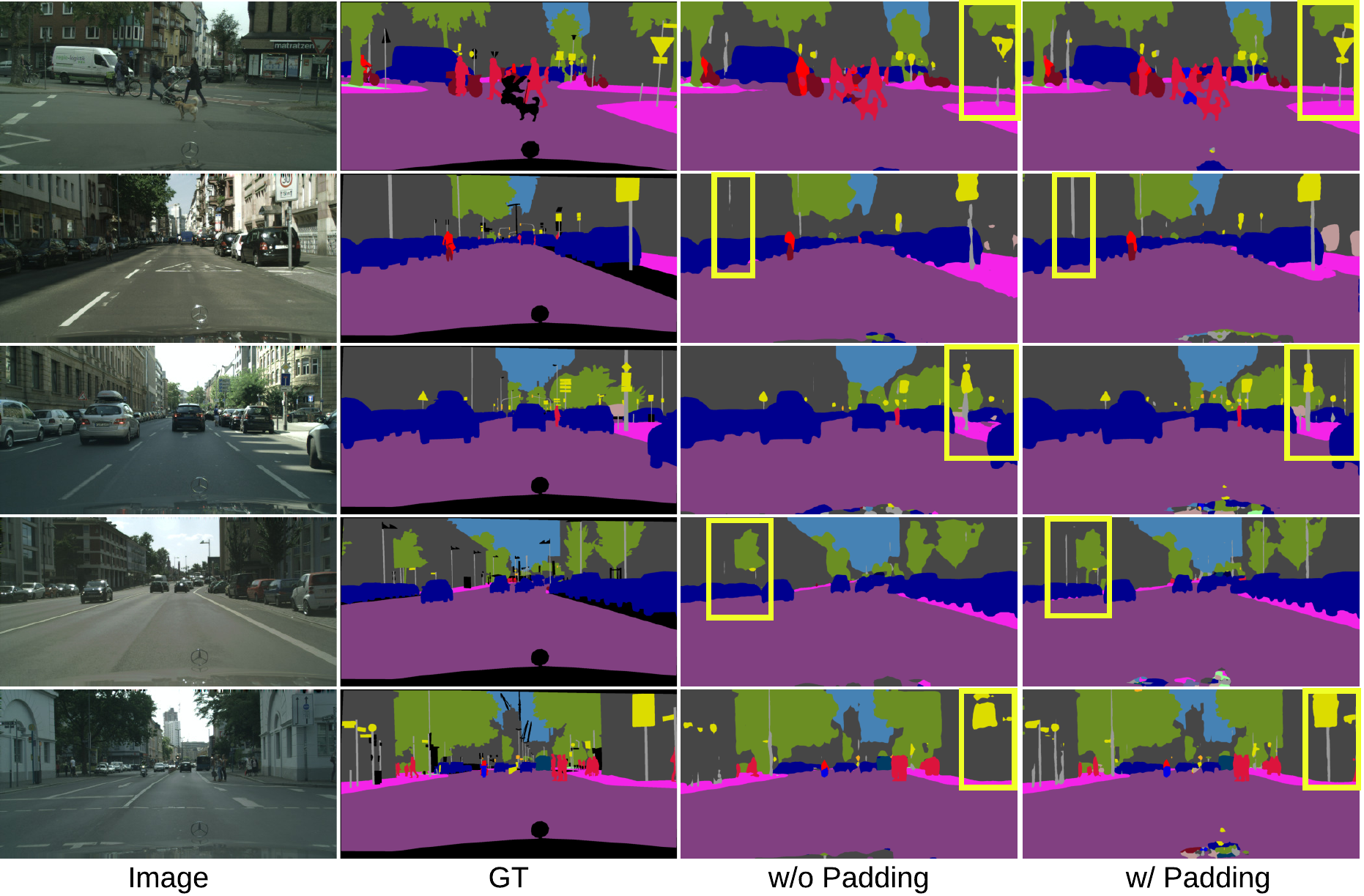}}
	\caption{\textbf{Example predictions of DeepLabv3-ResNet50 on the Cityscapes validation set} when training w/ and w/o \textit{padding} settings.}\label{fig:city_preds}
\end{figure*}
\subsection{Location Dependant Image Segmentation Predictions}\label{sec:sem_pred}
Figure~\ref{fig:semantic_preds} shows predictions of the location dependant image segmentation task for a grid size $k=5$. We visualize the predictions as a heatmap, where each pixel is colored according to the confidence that the semantic category appears in that pixel's location. We show predictions with padding (left) and without padding (right) for various grid locations, $L$. Note how boundary effects significantly impact locations near the border. In particular, locations in the corners are most affected, as they suffer from boundary effects originating from two borders (e.g., top and left border for $L=1$). 

Figure~\ref{fig:semantic_preds} shows predictions of the location dependant image segmentation task for a grid size $k=5$. We visualize the predictions as a heatmap, where each pixel is colored according to the confidence that the semantic category appears in that pixel's location. We show predictions with padding (left) and without padding (right) for various grid locations, $L$. Note how boundary effects significantly impact locations near the border. In particular, locations in the corners are most affected, as they suffer from boundary effects originating from two borders (e.g., top and left border for $L=1$). 
\begin{table}
\caption{\textbf{IoU comparison of DeepLabv3} for semantic segmentation task with three different \textit{padding} (Zero, Reflect, and No pad) settings.}
\centering 
      \begin{center}
         
      \setlength\tabcolsep{4.1pt}
		\def\arraystretch{1.2}
    \resizebox{0.39\textwidth}{!}{
         \begin{tabular}{c|ccc}
				 \specialrule{1.2pt}{1pt}{1pt}
				  Eval. Region &   Zero Pad & 
				  Reflect &No Pad\\
				  \specialrule{1.2pt}{1pt}{1pt}
				 0\%- 5\% & 72.6 &71.9& 63.7 \\
				 5\%- 10\% & 72.7 &72.0& 66.4\\
				 10\%- 15\% & 73.8 &73.7& 67.2\\
				 15\%- 20\% & 73.9 &74.1& 67.9\\
				 20\%- 25\% & 74.7 &74.8& 68.5\\
				 25\%- 30\% & 75.3 &75.4& 69.6\\
				 30\%- 35\% & 75.1 &75.2& 69.4\\
				 35\%- 40\% & 74.7 &75.2& 69.3\\
				 40\%- 45\% & 74.4 &74.8& 69.2\\
				 45\%- 50\% & 74.2 &74.5& 69.4\\
				 50\%- 55\% & 74.4 &74.9& 69.8\\
				 55\%- 60\% & 74.3 &74.8& 69.7\\
				 60\%- 65\% & 73.8 &74.3& 69.2\\
				 65\%- 70\% & 73.8 &74.4& 68.8\\
				 70\%- 75\% & 73.9 &74.5& 68.9\\
				 75\%- 80\% &  73.8 &74.4& 69.2\\
				 80\%- 85\% & 73.5 &74.1& 68.1\\
				 85\%- 90\% & 71.4 &71.9& 65.1\\
				 90\%- 95\% & 71.3 &72.0& 64.2\\
				 95\%- 100\% & 69.7 &70.1& 70.2\\
				 \hline
				 \textbf{Overall} & \textbf{74.0}& 73.9&\text{69.1}\\
				\specialrule{1.2pt}{1pt}{1pt}
			\end{tabular}
			}
			
			\label{tab:city_s}
      \end{center}
  \end{table}
  
  \begin{figure}
    \begin{center}
    \resizebox{0.44
    \textwidth}{!}{
          \includegraphics[width=0.44\textwidth]{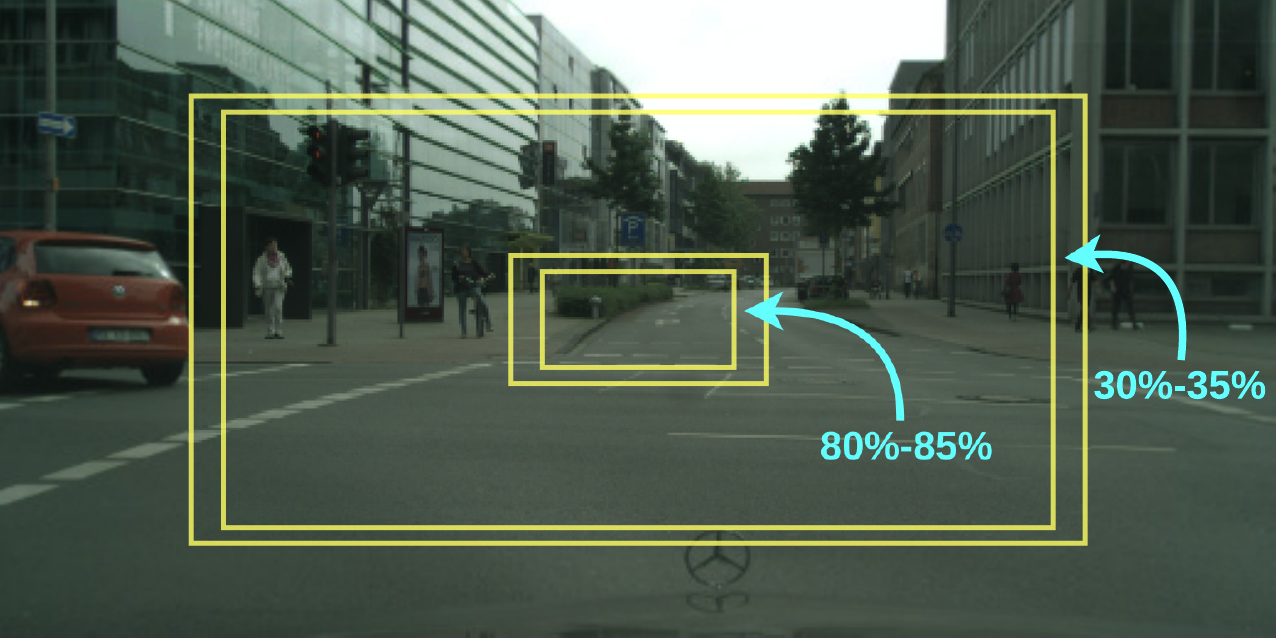}}
	\vspace{-0.2cm}
	\caption{An illustration of the evaluation regions used for the analysis in Table~\ref{tab:city_s} and Fig.~\ref{fig:city_supp}.} \label{fig:city_region}
	
	\vspace{0.5cm}
  \resizebox{0.49\textwidth}{!}{
  
    \begin{tikzpicture}\ref{test}
    \begin{axis}[
      line width=1.0,
        title style={at={(axis description cs:0.5,1.06)},anchor=north,font=\small},
        xlabel={Region Distribution},
        ylabel={mIoU(\%)},
        xmin=-5, xmax=105,
        ymin=60, ymax=76,
        xtick={10,20,30,40,50,60,70,80,90,100},
        ytick={65,70,75,80,85,90},
        x tick label style={font=\tiny},
        y tick label style={font=\tiny},
        xtick pos=bottom,
        x label style={at={(axis description cs:0.5,0.02)},anchor=north,font=\small},
        y label style={at={(axis description cs:0.1,.5)},anchor=south,font=\small},
        width=8.5cm,
        height=6.5cm,        
        ymajorgrids=false,
        xmajorgrids=false,
        grid style=dashed,
        legend style={
         nodes={scale=0.9, transform shape},
         cells={anchor=west},
         legend style={at={(3.6, 0.65)},anchor=south}},
         legend columns=1,
         legend entries={[black]w/o padding, [black]Reflect padding, [black]Zero padding},
         legend to name=test,
    ]
    \addplot[line width=1.2pt,mark size=1.7pt,smooth,color=blue,mark=*,]
        coordinates {(5,63.7)(10,66.4)(15,67.2)(20,67.9)(25,68.5)(30,69.6)(35,69.4)(40,69.3)(45,69.2)(50,69.4)(55,69.8)(60,69.7)(65,69.2)(70,68.8)(75,68.9)(80,69.2)(85,68.1)(90,65.1)(95,64.2)(100,61.2)};
        
    \addplot[line width=1.2pt,mark size=1.7pt,smooth,color=red,mark=*,]
        coordinates {(5,71.9)(10,72.0)(15,73.7)(20,74.1)(25,74.8)(30,75.4)(35,75.2)(40,75.2)(45,74.8)(50,74.5)(55,74.9)(60,74.8)(65,74.3)(70,74.4)(75,74.5)(80,74.4)(85,74.1)(90,71.9)(95,72.0)(100,70.1)};
        
    \addplot[line width=1.2pt,mark size=1.7pt,smooth,color=orange,mark=*,]
        coordinates {(5,72.6)(10,72.7)(15,73.8)(20,73.9)(25,74.7)(30,75.3)(35,75.1)(40,74.7)(45,74.4)(50,74.2)(55,74.4)(60,74.3)(65,73.8)(70,73.8)(75,73.9)(80,73.8)(85,73.5)(90,71.4)(95,71.3)(100,69.7)};
    \end{axis}
    \end{tikzpicture} }
    \vspace{-0.5cm}
  \caption{Performance comparison of DeepLabv3 network with respect to various image regions and padding settings used in Table~\ref{tab:city_s}.}
  \label{fig:city_supp}
  \end{center}
\end{figure} 

\subsection{Extended Boundary Effect Analysis on Cityscapes Dataset}\label{sec:city_supp}
\vspace{-0.2cm}
We continue to investigate the impact that zero padding has on the ability of a strong and deep CNN to segment objects near the image boundary. Results shown use the same network and training settings as in Sec. 7 of the main manuscript, on the Cityscapes~\cite{cordts2016cityscapes} dataset. We first show additional qualitative examples in Fig.~\ref{fig:city_preds}, which clearly shows a large reduction in performance at locations near the border when no padding is used, particularly for thin objects (e.g., street lamps or column poles). 

We present additional results (see Table~\ref{tab:city_s} and  Fig.~\ref{fig:city_supp}) of the analysis presented in Sec. 6 (semantic segmentation) in the main paper. 
Fig.~\ref{fig:city_region} shows sample evaluation regions used for this analysis. The no padding case has a steeper drop-off in performance as regions of evaluation get closer to the image boundary. Note how, in all cases, the performance increases from the border to the inner $25\%$, at which point the performance is somewhat stagnant until it reaches the innermost $80\%$. 

Surprisingly, we also observe a \textit{steeper} drop off in the middle of the image for the no padding case, supporting our hypothesis that boundary effects play a role at all regions of the image without the use of padding. We believe the drop in performance at the center regions is due to Cityscapes being an automotive-centric dataset, where pixels at the center of the image are often at large distances away from the camera, unless the vehicle collecting the data has an object directly in front of it.

\subsection{Canvas Analysis: Cutout \& Adversarial Robustness}
\begin{figure}
	\begin{center}
		\includegraphics[width=0.44\textwidth]{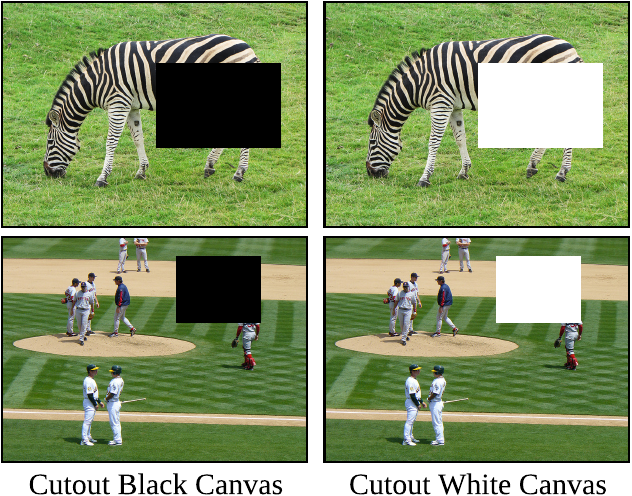}
		\caption{Sample training images generated using Cutout~\cite{devries2017improved} under two different canvases.}
		\label{fig:cutout}
	\end{center}
\end{figure}

Figure~\ref{fig:cutout} shows two training examples of Cutout strategy. Following Cutout, we simply place a rectangular mask (black and white) over a random region during the training. Note that we evaluate on the standard PASCAL VOC 2012 validation images. 

\end{document}